\documentclass{article}
\usepackage{fullpage} 

\usepackage[titletoc, title]{appendix}
\usepackage{graphicx}
\graphicspath{{fig/}}
\usepackage{caption}
\usepackage{subcaption}

\usepackage{float}

\usepackage{multirow}

\usepackage{natbib}

\usepackage{amsmath, amssymb, amsthm}

\newcommand{\cH}{\mathcal{H}}

\newcommand{\Rank}{\mathrm{rank}}

\title{An Empirical Comparison of Sampling Quality Metrics:\\
 A Case Study for Bayesian Nonnegative Matrix Factorization}

\author{Arjumand Masood$^*$ \\
 Harvard University
 \and Weiwei Pan$^*$\\
 Harvard University
 \and Finale Doshi-Velez \\
Harvard University}
\begin{document}

\maketitle

\section{Introduction}
Bayesian approaches to machine learning begin by positing that the
data $X$ can be explained by some probablistic model $p(X|\theta)$,
where $\theta$ is a set of parameters.  Rather than finding a point
estimate for $\theta$ that maximizes the likelihood $p(X|\theta)$,
Bayesian approaches place a a prior distribution over the parameters
$p(\theta)$ and compute the posterior $p(\theta|X)$.  The posterior
$p(\theta|X)$ captures uncertainty in the parameters $\theta$.  

For most models, the posterior $p(\theta|X)$ does not have an analytic
form.  In this situation, a popular approach is to approximate the
posterior $p(\theta|X)$ through a set of samples
$\{\theta_1,..,\theta_N\}$.  Approaches for generating these samples
include importance and rejection sampling \cite{liu1996metropolized}, sequential Monte
Carlo \cite{halton1962sequential}, and Markov Chain Monte Carlo \cite{gilks2005markov}.\footnote{There are
  other approaches for approximating posterior distributions, such as
  variational methods \cite{wainwright2008graphical, tzikas2008variational, opper2009variational}; we focus on sampling-based methods here
  but the ideas are generally applicable.} 

In this work, we empirically explore the question: how can we assess
the quality of these samples?  We assume that the samples are provided
by some valid Monte Carlo procedure, so we are guaranteed that the
collection of samples will asymptotically approximate the true
posterior $p(\theta|X)$.  Most current evaluation approaches focus on
two questions: (1) Has the chain mixed, that is, is it sampling from
the posterior $p(\theta|X)$? and (2) How independent are the samples
(as MCMC procedures produce correlated samples)?  
Focusing on the case of Bayesian nonnegative matrix factorization, 
we empirically evaluate standard metrics of sampler quality as well as 
propose new metrics to capture aspects that these measures fail to expose.
The aspect of sampling that is of particular interest to us is the ability (or inability) of
sampling methods to move between multiple optima in NMF problems. As a proxy, 
we propose and study a number of metrics that might quantify the diversity of a set 
of NMF factorizations obtained by a sampler through quantifying the coverage of 
the posterior distribution. We compare the performance of a number of standard 
sampling methods for NMF in terms of these new metrics.


\section{Background} 

\subsection{Current Measures of Sampling Quality} \label{sec:background}
\paragraph{Measures of Mixing.} 
While it is practically impossible to assess whether a chain has
mixed, some popular approaches include sample paths, cumulative sums, autocorrelation plots, batch means, AR and spectral analysis estimators \cite{johnson1996studying, cowles1996markov, flegal2010batch}. 

\paragraph{Measures of Sample Independence.} 
Most current approaches to measuring sampling quality focus on measuring the
independence the samples from the chain.  Since sample independence is hard to
assess, most measures focus on correlation between samples. These include: effective sample size, 
autocorrelation plots, cross-correlation, integrated autocorrelation time, Hairiness Index \cite{cowles1996markov, brooks1998quantitative}. 

\paragraph{Other Measures} 
In the theoretical literature, other popular measures include cover
time, hitting time, etc \cite{aldous2002reversible, lee2015efficiency}. However, these are impractical in large scenarios or when the modes of the posterior distribution are unknown.

\subsection{Bayesian Non-negative Matrix Factorization}\label{sec:nmfback}
In the case study below, we will evaluate several metrics of sample
quality in the context of Bayesian nonnegative matrix factorization.
We choose this example because it is one of the simplest and popular
data exploration techiniques---NMF has been used to in wide-ranging
applications ranging from understanding protein-protein interactions
\cite{Greene}, finding topics in large text corpora
\cite{roberts2016navigating}, and discovering molecular pathways from
genomic samples \cite{brunet2004metagenes}---with a myriad of
efficient algorithms for solving it
\cite{paisley2015bayesian,schmidt2009bayesian,moussaoui2006separation,lin2007projected,lee2001algorithms,recht2012factoring}.
However, NMF still suffers from non-identifiability: even in the exact
case, there can be multiple solutions.  Thus, it serves a good test
case for measuring how well current Bayesian approaches can describe this
uncertainty.  

\paragraph{Nonnegative Matrix Factorization and Identifiability} 
Given an $D \times N$ nonnegative matrix $X$ and desired rank $R$, the
nonnegative matrix factorization (NMF) problem involves finding an $R
\times N$ nonnegative weight matrix $W$, and an $D \times R$
nonnegative basis matrix $A$, such that $X \approx AW$.  The ease of
interpreting the weights $W$ and bases $A$ (due to the nonnegativity
constraints), and the myriad of efficient algorithms for solving NMF
\cite{paisley2015bayesian,schmidt2009bayesian,moussaoui2006separation,lin2007projected,lee2001algorithms,recht2012factoring},
has made NMF---and related models, such as topic models---a popular
approach to data exploration in many fields.  NMF has been used to
understand protein-protein interactions \cite{Greene}, find topics in
large text corpora \cite{roberts2016navigating}, and discover
molecular pathways from genomic samples \cite{brunet2004metagenes}.

However, in many cases NMF is not identifiable: there may be very
different pairs $(A,W)$ and $(A',W')$ that might explain the data $X$
(perhaps almost) as well.  In the following, we briefly recall some relevant
terminology and properties of NMF related to the notion of identifiability. 

If $X = AW$, we call the pair $(A, W)$ an \emph{exact NMF}; the
minimum rank $R$ such that $X$ admits an exact NMF is called the
\emph{nonnegative rank of $X$} and is denoted $\Rank_+(X)$.  
A nonnegative matrix factorization, $X=AW$, can be considered
trivially non-unique.  Given any permutation matrix $P$ and diagonal
matrix $D$ with positive entries, we obtain an alternate factorization
of $X$, namely, $X = (AP^\top D^{-1})(DPW)$. Since the factorization
($AP^\top D^{-1}$, $DPW$) differs from ($A$, $W$) by scaling and
relabeling of the column vectors in $A$, we consider them
equivalent. Thus, we call an NMF \emph{unique} if all solutions can be
represented as $AQQ^{-1}W$, where $Q$ is a monomial matrix (i.e. a
product of some $P$ and some $D$).

Awareness and concerns of non-identifabiltity has been gaining
attention among practitioners. For example, \cite{Greene} use
ensembles of NMF solutions to model chemical interactions, while
\cite{roberts2016navigating} conduct a detailed empirical study of
multiple optima in the context of extracting topics from large
corpora.  These approaches use random restarts to find multiple
optima.

\paragraph{Bayesian Nonnegative Matrix Factorization} 
Bayesian NMF approaches \cite{schmidt2009bayesian,
  moussaoui2006separation} promise to characterize parameter
uncertainty in a principled manner by solving for the posterior
$p(A,W|X)$ given priors $p(A)$ and $p(W)$.  Having such a
representation of uncertainty in the bases and weights can further
assist with the proper interpretation of the factors: we may place
more confidence in subspace directions with low uncertainty, while
subspace directions with more uncertainty may require further
exploration.  Unfortunately, in practice, the uncertainty estimates
are often of limited use: sampling-based approaches
\cite{schmidt2009bayesian, moussaoui2006separation} rarely switch
between multiple modes.

In this case study, we will use the generative model of Schmidt et. al. in 
\cite{schmidt2009bayesian}. In this case, we place exponential priors $p(A)$ and $p(W)$ on $A$ and $W$ and choose 
a Gaussian likelihood. Thus, the elements of $W$, $A$ are sampled from a
rectified normal distributions $\mathcal{R}(x; \mu,\sigma^2,\lambda)$,
which is proportional to the product of a Gaussian and an exponential
$\mathcal{N}(x; \mu,\sigma^2)\text{Exp}(x; \lambda) $. The full
condition for the entries in $A|Q$ is given by
\begin{align}
\nonumber p(A_{d,r} &| X, A_{\backslash (d,r)}, W, \sigma^2) = \mathcal{R}(A_{d,r}; \mu_{A_{d,r}},\sigma^2_{A_{d,r}},\lambda_{A_{d,r}}) \\ 
\nonumber \mu_{A_{d,r}} &= \frac{\sum_n (X_{d,n} - \sum_{r' \neq r} A_{d,r'} W_{r',n})}{\sum_n W_{r,n}^2},\quad \sigma^2_{A_{d,r}} = \frac{\sigma^2}{\sum_n W_{r,n}^2}
\end{align}
with a symmetric update for $W$.

\section{Additional Measures of Sample Quality: Notions of Coverage} 
The measures in Section \ref{sec:background} largely focus on the independence
of samples.  However, even in relatively simple models, such as
Bayesian NMF, it is possible for a chain to mix quickly within a
single mode and never reach an alternate mode or region.  What is
often missing in our discussion of practical MCMC approaches is a
notion of coverage: In addition to moving in ``independent" ways, how
much of the posterior space does a finite-length chain explore?  
(Obviously given infinite time, every correct MCMC procedure will find all the modes.)  
In this section, we describe several easy-to-compute and principled measures
of coverage that can be applied to any set of samples, whether or not
they come from a Markov Chain.  

\subsection{Measures of Similarity}
To quantify the ``diversity" of a set of samples , we first need a
notion of distance or similarity between a pair of samples.  Below, we
describe two matrix similarity measures, which can be meaningfully
interpreted in the context of Bayesian NMF. Later, we show that the
choice of one similarity measure may be more appropriate than another
depending on the NMF model and the application.

Recall from Section \ref{sec:nmfback} that we consider two basis matrices
$A$ and $A'$ to be equivalent (defining the same factorization of $X$)
when $A' = AQ$ for some monomial matrix $Q$. Thus, we need to ensure
that each similarity measure we construct is defined on equivalence
classes of matrices; that is, the disimilarity of two matrices in the
same class should be zero. To do this, we scale each column in our
matrices to be unit in some norm and we use permutation invariant representations
of matrices (e.g. matrices as unordered collections of column vectors).
 
\paragraph{Minimum Matching Distance}
For a fixed metric $m$ on $\mathbb{R}^D$, the minimum matching distance is a metric supported on sets of vectors in $\mathbb{R}^D$ \cite{walters2011random}. Given two subsets of $\mathbb{R}^D$, $A=\{A_1, \ldots, A_R\}$ and $A' = \{A'_1, \ldots, A'_R \}$, their minimum matching distance is defined as
\begin{align}
d_{\mathrm{MM}}(A, A') =  \min_{\sigma \in S_R} \sum_{r=1}^R d(A_{\sigma(r)}, A'_{\sigma(r)})
\end{align}
where $S_R$ is the set of of permutations of the index set $\{1, \ldots, R\}$. Intuitively, the minimum matching distance measures the total distance of the corresponding vectors of $A$ in $A'$, minimized over all bipartite matchings of the vectors. The minimum matching distance can be efficiently computed using the Kuhn-Munkres algorithm \cite{kriegel2003using, brecheisenefficient}.

\paragraph{$\ell_1$ Matching}
We fix the metric as $\ell_1$. Given $A, A' \in \mathbb{R}^{R\times D}$ each with  columns that are unit $\ell_1$-norm, we can compute their minimum matching distance as:
\begin{equation}
d_{\mathrm{MM}_{\ell_1}}(A,A') = \min_{P \in \mathcal{P}}{ \| AP - A' \| _1},
\end{equation}
where $\mathcal{P}$ is the set of $R \times R$ permutation matrices. Note that since the columns of $A$ and $A'$ are unit $\ell_1$-norm, $d_{\mathrm{MM}_{\ell_1}}(A, A')$ is bounded between $0$ and $R$.

The $\ell_1$ minimum matching distance measurement is closely related to the total variation distance for comparing discrete probability distributions. In the case of topic modeling, each column of $A$ or $A'$ corresponds to a ``topic", which can be interpreted as a probability distribution over words in a dictionary. The $\ell_1$ minimum matching distance is (up to scale) the total variation distance after pairwise matching topics in $A$ and $A'$ based on similarity. 

\paragraph{Maximum Angle Similarity}
Masood et al \cite{Masood} defines an angle-based similarity measurement that aims to capture the permutation ambiguity as well as the essence of a diverse factorization. Given $A, A' \in \mathbb{R}^{R\times D}$, let $\widehat{\sigma}$ be a permutation of the columns of $A'$ that minimizes the average angle between corresponding columns,
\begin{align}
\widehat{\sigma} = \arg\min_{\sigma\in S_R} \frac{1}{R}\sum_{r \in R} \cos^{-1}\left(\frac{A_r \cdot A'_{\sigma(r)}}{\|A_r\|\|A'_{\sigma(r)}\|}\right)
\end{align}
The maximum angle similarity of $A$ and $A'$ is defined largest angle between corresponding columns, under the permutation $\widehat{\sigma}$,
\begin{equation*}
d_{\text{Angle}}(A,A') = \max_{r \in R}\; \cos^{-1}\left(\frac{A_r \cdot A'_{\widehat{\sigma}(r)}}{\|A_r\|\|A'_{\widehat{\sigma}(r)}\|}\right)
\end{equation*}
As the entries in $A$ and $A'$ are non-negative, the above dot product must always be non-negative. Thus, $d_{\text{Angle}}(A,A')$ is bounded between $0$ and $\pi/2$.

Since the maximum angle similarity is measurement of orientation, we can apply the same measurement to basis elements of different dimensions and interpret the results in a similar manner. By focusing only on measuring the maximum angle, we allow for an extreme case where two factorizations only differ by one basis element. This may well be the case in some scenario so we incorporate it as a feature of this distance measurement. 

In addition, by considering the maximum angle, our similarity measure gains a certain degree of robustness to column perturbations. For example, a similarity  measurement based on the sum of angles between basis elements would not be successful in differentiating between a case where one basis element is significantly different versus if all basis elements were just perturbed a little bit. We want our understanding of diversity to be robust to perturbation of existing basis columns. 

\begin{figure}[H]
    \centering
    \begin{subfigure}[b]{0.44\textwidth}
        \centering
        \includegraphics[height=35mm]{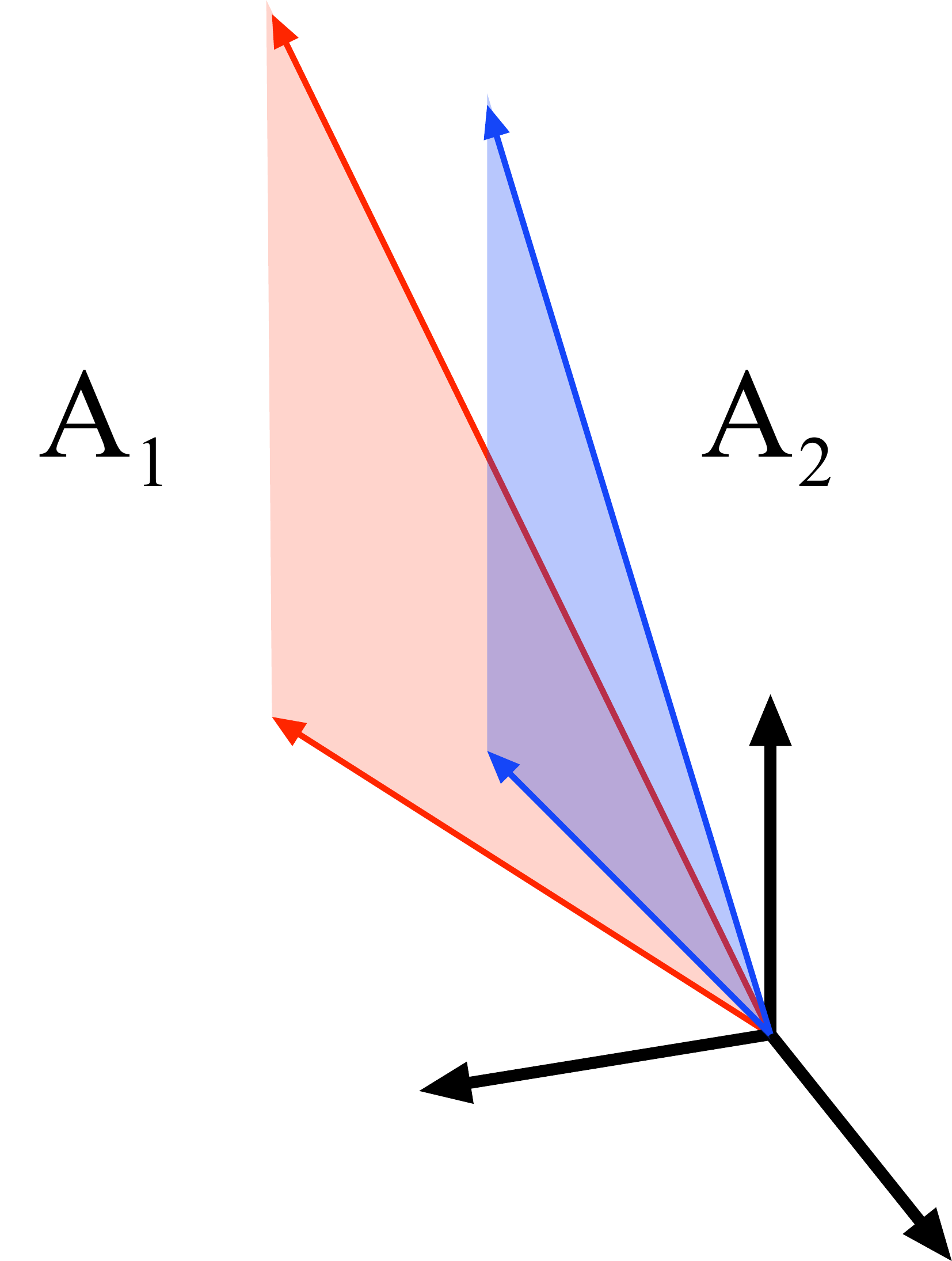}
        \caption{$d_{\mathrm{MM}_{\ell_1}}(A_1,A_2) =  0.137$,  $d_{\text{Angle}}(A_1,A_2) = 6.38$}
    \end{subfigure}    \hskip1cm
    \begin{subfigure}[b]{0.44\textwidth}
        \centering
        \includegraphics[height=35mm]{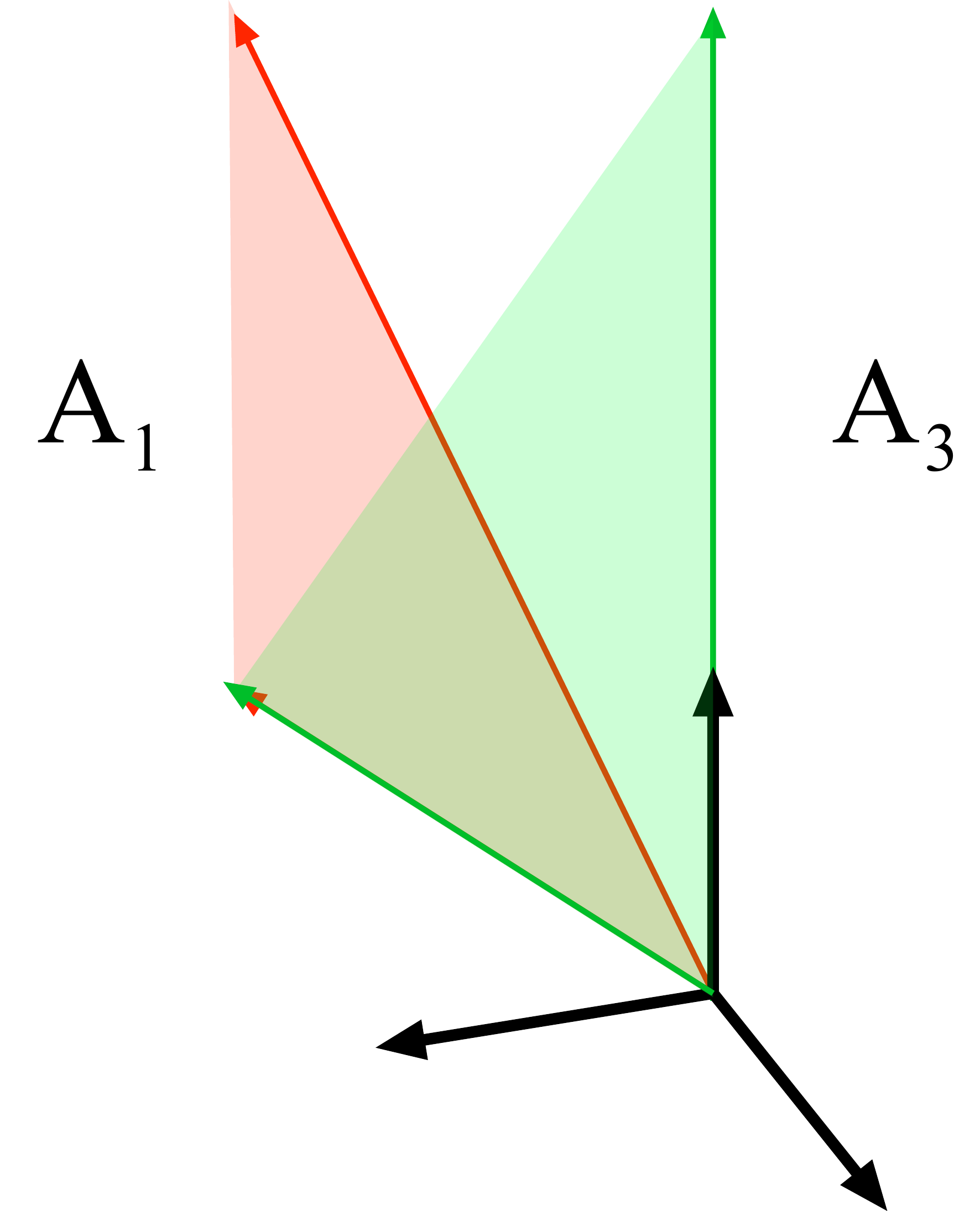}
        \caption{$d_{\mathrm{MM}_{\ell_1}}(A_1,A_3) =  0.400$,  $d_{\text{Angle}}(A_1,A_3) = 20.44$}
    \end{subfigure}   
    \caption{Comparison of max angle and $\ell_1$-min match measures when $\|A_1 - A_2 \|_F = \| A_1 - A_3 \|_F$. }
     \label{fig:As}
\end{figure}

\paragraph {Notes and Observations about similarity in NMF basis factors:} Consider as a baseline, measuring the similarity as the Frobenius error of the difference i.e. $d_F(A,A') = \| A - A'\|_F$. Figure \ref{fig:As} shows two pairs of matrices $(A_1, A_2)$ and $(A_1, A_3)$ that have the same difference in the naive Frobenius sense ($d_F$ = 0.122) but different maximum angle and $l_1$-minimum matching similarity scores. From figure \ref{fig:As} it is also clear that the $(A_1,A_3)$ pair is the more dissimilar. This example illustrate the fact that these measures of similarity indeed capture the sense of diversity in solutions that is of interest to us.


Finally, we point out that the permutation computed in the minimum matching distance, which minimizing the total distance between matrices, is not necessarily the same as the permutation, which minimizes the average angle between corresponding columns. In practice, this means that permutations used to compute the minimum matching distance cannot be used to compute the maximum angle distance (and vice versa). It also raises an interesting question about what is means in empirical work to `correct' for the permutation ambiguity since we've observed that this correction is metric-dependent. 

\subsection{Measures of Coverage}
Given a set, $S$, of samples from some parameter space explore by a sampler, and given a similarity measure $m$ for pairs of samples, we introduce three measurement of the diversity of the samples contained in $S$.

\paragraph{Maximum Pairwise Distance}
We define a notion of diversity for a set $S$ based on the ``diameter" of $S$, that is, we compute the maximum distance between pairs of points points in $S$. The maximum pairwise distance of $S$ is defined as 
\begin{align}
\mathrm{MaxDist}(S) = \max_{A, A' \in S} d(A, A).
\end{align}

\paragraph{Mean Pairwise Distance}
We can alternatively quantify the diversity of $S$ by approximating the ``density" of points in $S$. Motivate by this intuition, we define the mean pairwise distance of $S$ as 
\begin{align}
\mathrm{MeanDist}(S) = \frac{1}{|A|^2}\sum_{A, A' \in S} d(A, A).
\end{align}

\paragraph{Covering Number}
We quantify the amount of the parameter space explored by the sampler, by approximating a notion of``volume" for a set of samples $S$. 
For each $\epsilon>0$, we define the \emph{minimum covering number of $S$}, denoted $C_\epsilon(S)$, as the cardinality of the smallest subset $S'\subset S$ such that $\underset{s\in S'}{\bigcup}B(s, \epsilon)$ covers $S$, where $B(s, \epsilon)$ is the $\epsilon$ ball centered at $s$ with respect to some metric or similarity measure. Our minimum covering number can stated in terms of the covering number of graphs of the $\epsilon$-neighbor graph of the points in $S$. We note that the covering number is a frequently studied property of graphs in literature \cite{abbott1979bounds, chepoi2007covering}.

Clearly, $C_\epsilon(S)$ depends on the choice of $\epsilon$. When $\epsilon = 0$, the minimum covering number is equal to the cardinality of $S$; for sufficiently large $\epsilon$, the minimum covering number is 1, since an $\epsilon$-ball centered at any element will contain the entire set $S$. It is also straightforward to see that $C_\epsilon$ is a monotone increasing as a function of $\epsilon$. 

\begin{figure}[H]
    \centering
    \begin{subfigure}[b]{0.43\textwidth}
        \centering
        \includegraphics[height=30mm]{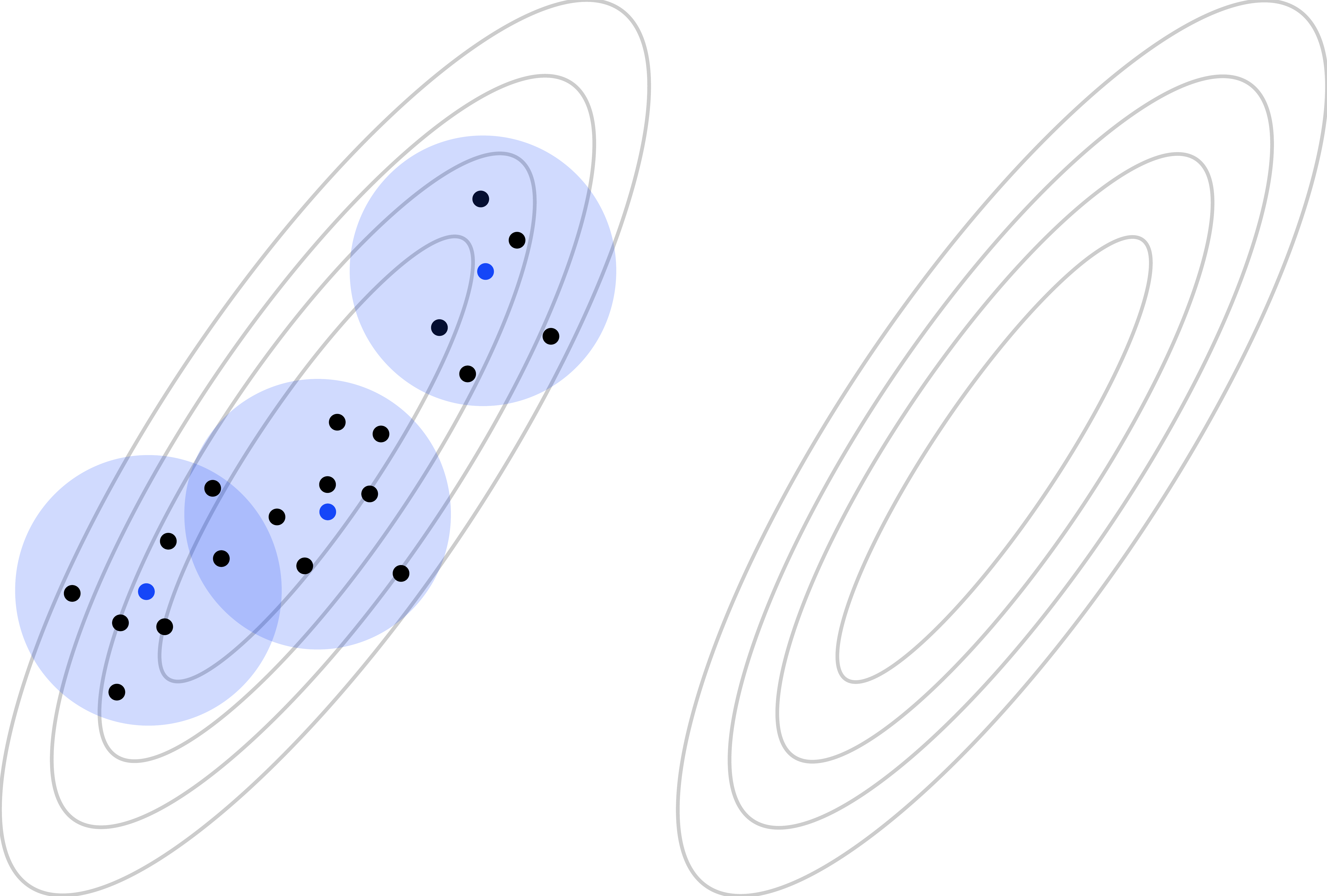}\hfill
        \caption{Sample $S_1$ lie in a single mode. The minimum covering number for $\epsilon=1$ is $C_\epsilon(S_1)=3$. }
    \end{subfigure} \hskip0.6cm     
    \begin{subfigure}[b]{0.43\textwidth}
        \centering
        \includegraphics[height=30mm]{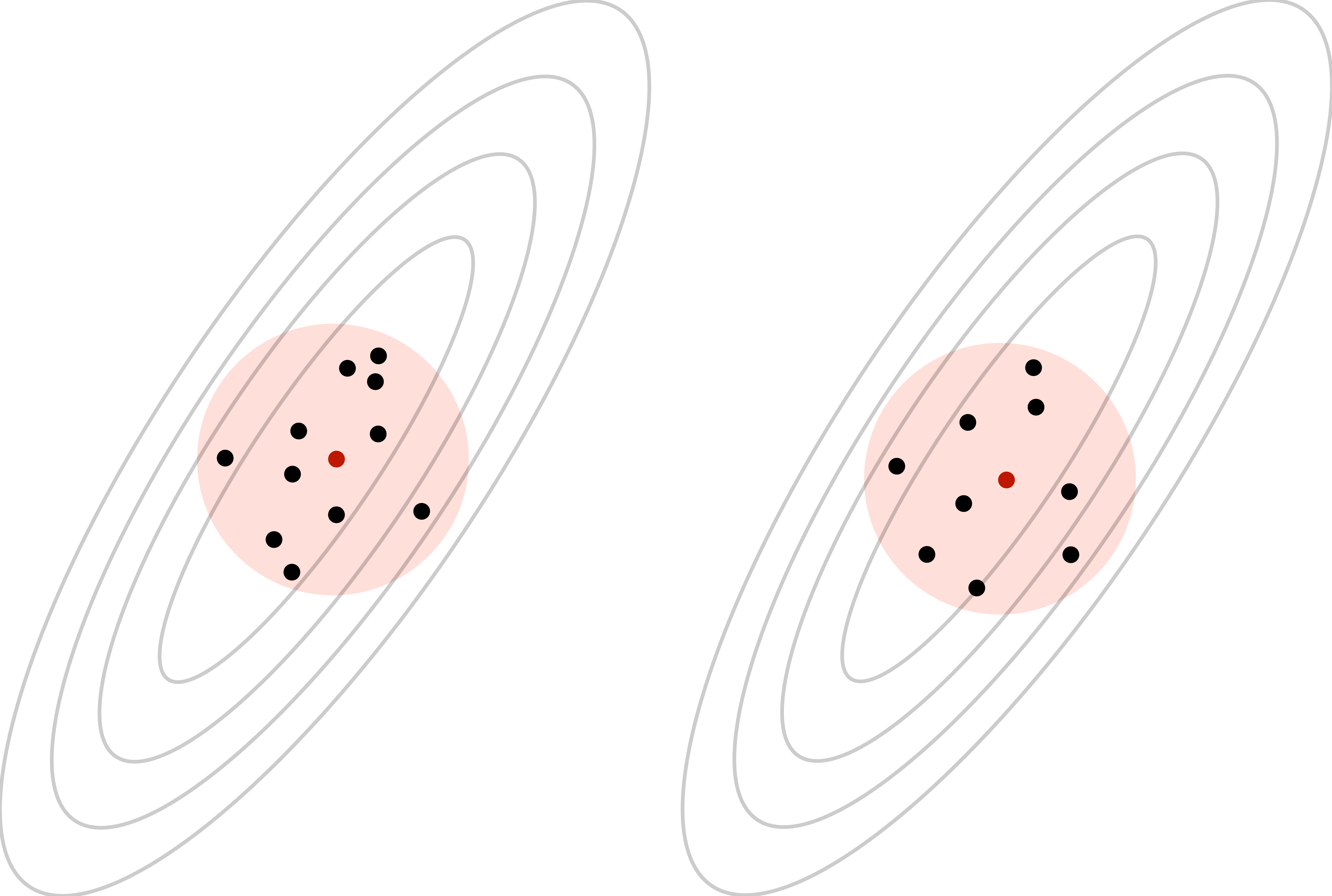}
        \caption{Sample $S_2$ lie in two modes. The minimum covering number for $\epsilon=1$ is $C_\epsilon(S_2)=2$. }
    \end{subfigure}   
    \caption{Comparison of minimum covering numbers of samples ($\epsilon=1$)}
    \label{fig:cov1}
\end{figure}

\begin{figure}[H]
    \centering
    \begin{subfigure}[b]{0.43\textwidth}
        \centering
        \includegraphics[height=30mm]{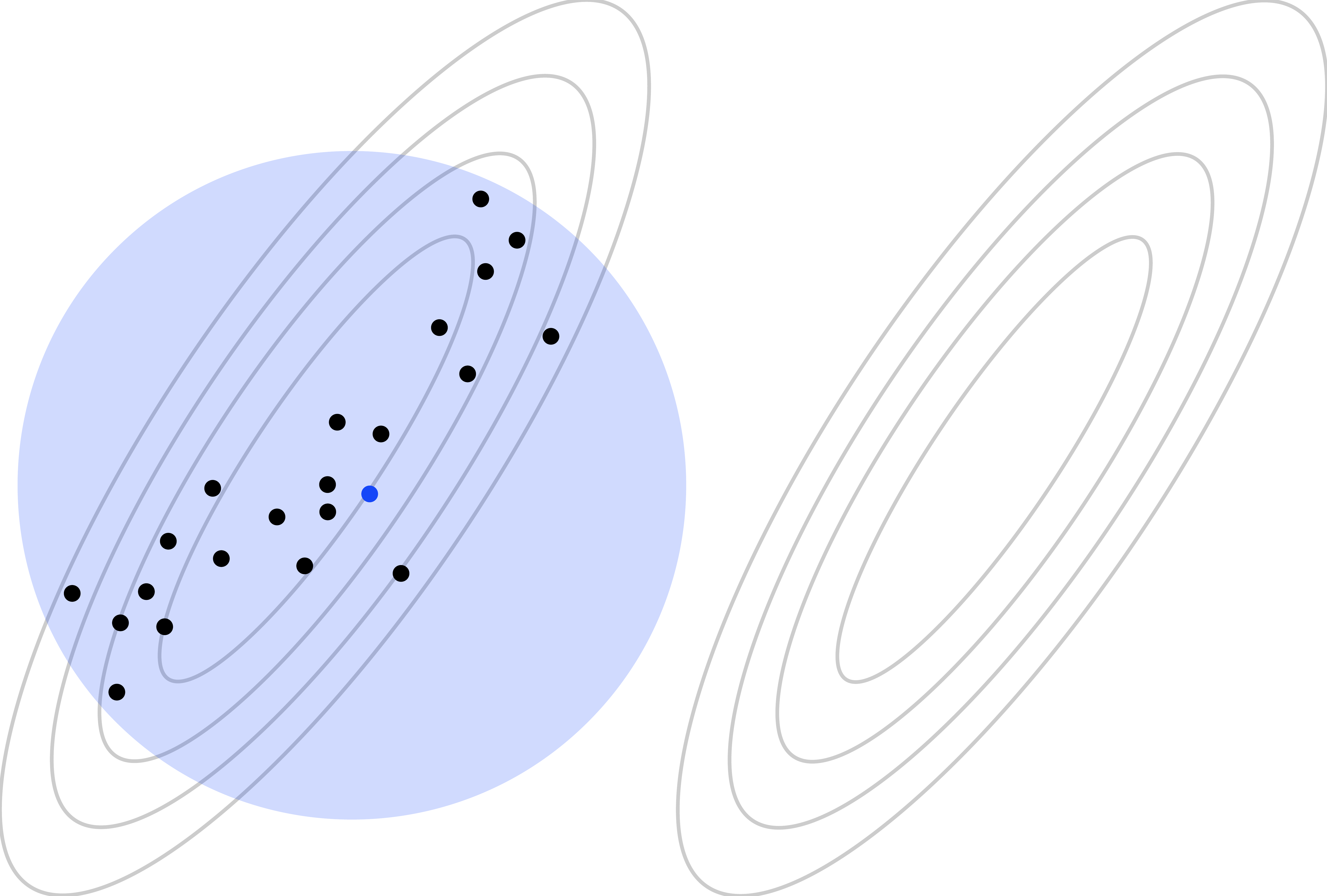}
        \caption{Sample $S_1$ lie in a single mode. The minimum covering number for $\epsilon=2.5$ is $C_\epsilon(S_1)=1$. }
    \end{subfigure} \hskip0.6cm     
    \begin{subfigure}[b]{0.43\textwidth}
        \centering
        \includegraphics[height=30mm]{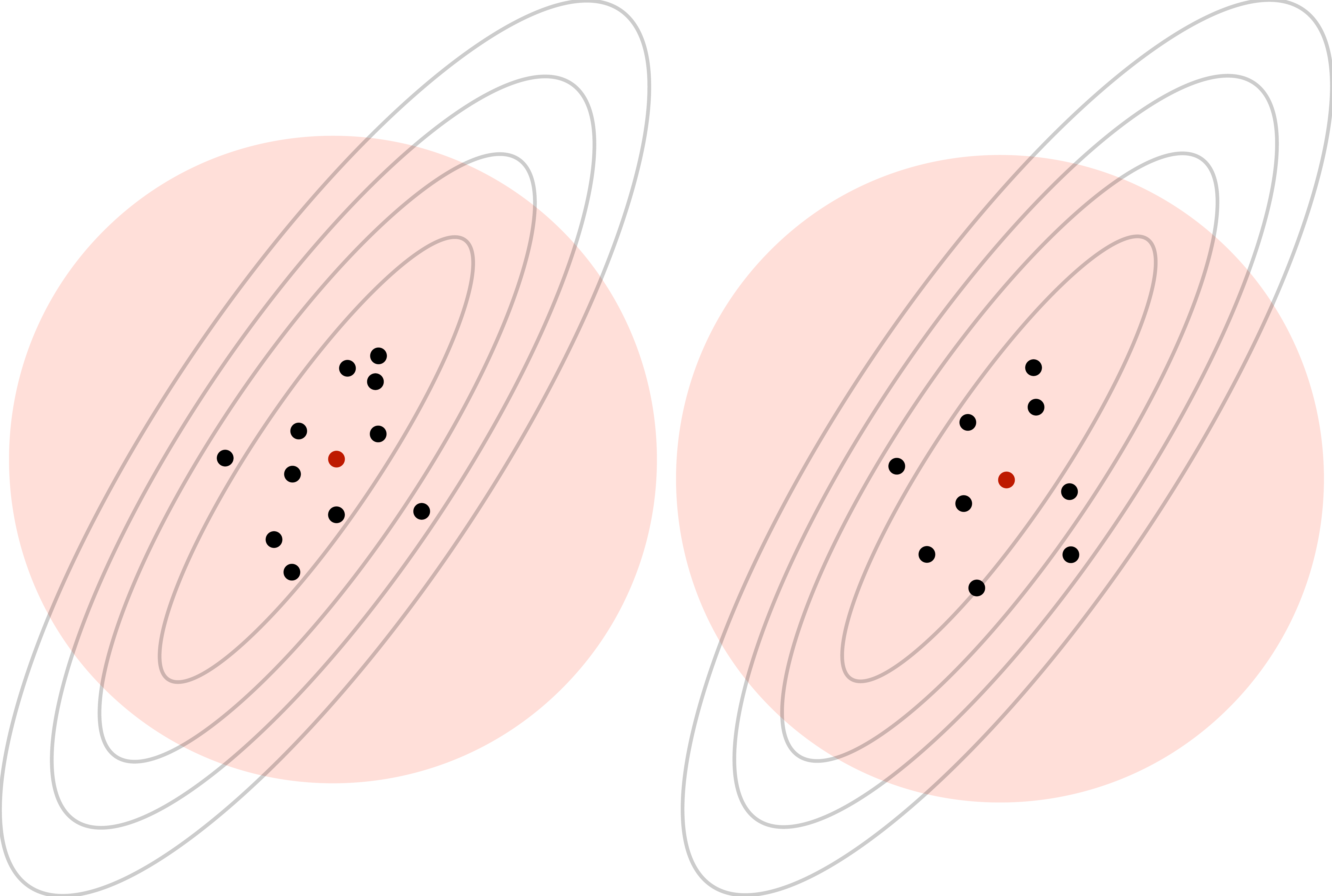}
        \caption{Sample $S_2$ lie in two modes. The minimum covering number for $\epsilon=2.5$ is $C_\epsilon(S_2)=2$. }
    \end{subfigure}   
    \caption{Comparison of minimum covering numbers of samples ($\epsilon=2.5$)}
     \label{fig:cov2}
\end{figure}

Generally speaking, for a fixed $\epsilon$, the larger the minimum covering number, the more of the parameter space covered by the sampler. However, cases may arise, for $\epsilon_1 < \epsilon_2$ and sets $S_1$, $S_2$, where $C_{\epsilon_1}(S_1) > C_{\epsilon_1}(S_2)$ but $C_{\epsilon_2}(S_1) < C_{\epsilon_2}(S_2)$. In Figure \ref{fig:cov1} and Figure \ref{fig:cov2}, we see samples $S_1$, $S_2$ from a mixture of two Gaussians, where $S_1$ is concentrated in one mode and $S_2$ is distributed amongst both modes. The latter is demonstrated by the fact that $C_{\epsilon_2}(S_2)> C_{\epsilon_2}(S_1)$ for sufficiently large choices of $\epsilon$. 

To avoid arbitrariness in selecting a single value for $\epsilon$, we compute the minimum covering number for a range for values between $\epsilon = 0$ and $\epsilon = t$, for $t$ is sufficiently large that each $\epsilon$-ball covers $S$. We interpret covering numbers which ``persists" for significant intervals of $\epsilon$ to be revealing the diversity of the sample set and covering numbers which appear for small intervals of $\epsilon$ to be negligible (due to small variations in the set). The motivation for our interpretation is based in the body of work on persistent topology \cite{edelsbrunner2008persistent, carlsson2009topology, ghrist2008barcodes}, in which topological features of manifolds are deduced from an approximation given by a set of sample points by interpreting the features which persist in the reconstruction across a number of resolutions.

\begin{figure}[H]
    \centering
    \begin{subfigure}[b]{0.43\textwidth}
        \centering
        \includegraphics[height=45mm]{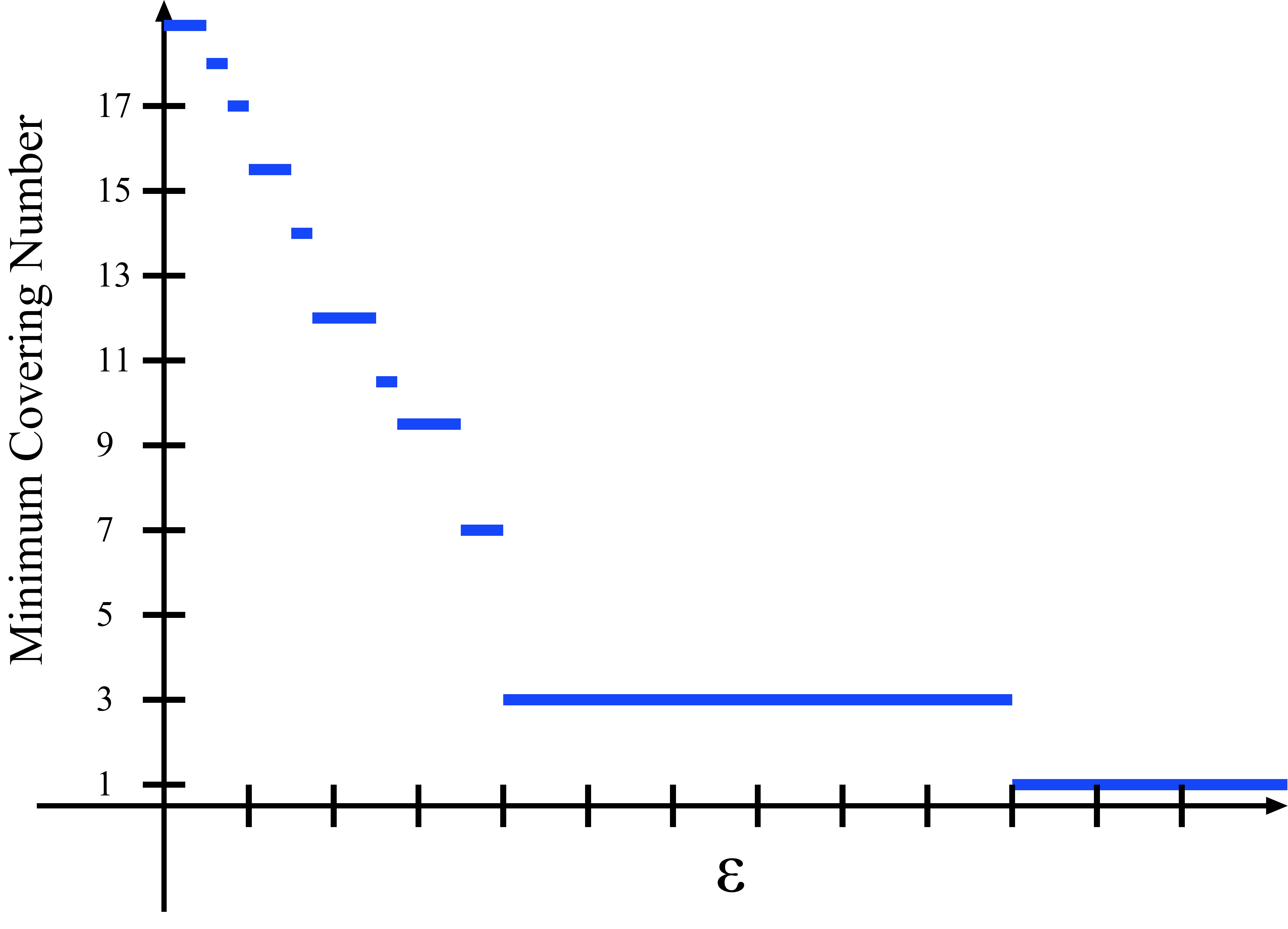}
        \caption{The minimum covering number of $S_1$ persists at 3 for 6 $\Delta\epsilon$'s.}
    \end{subfigure}    \hskip0.6cm   
    \begin{subfigure}[b]{0.43\textwidth}
        \centering
        \includegraphics[height=45mm]{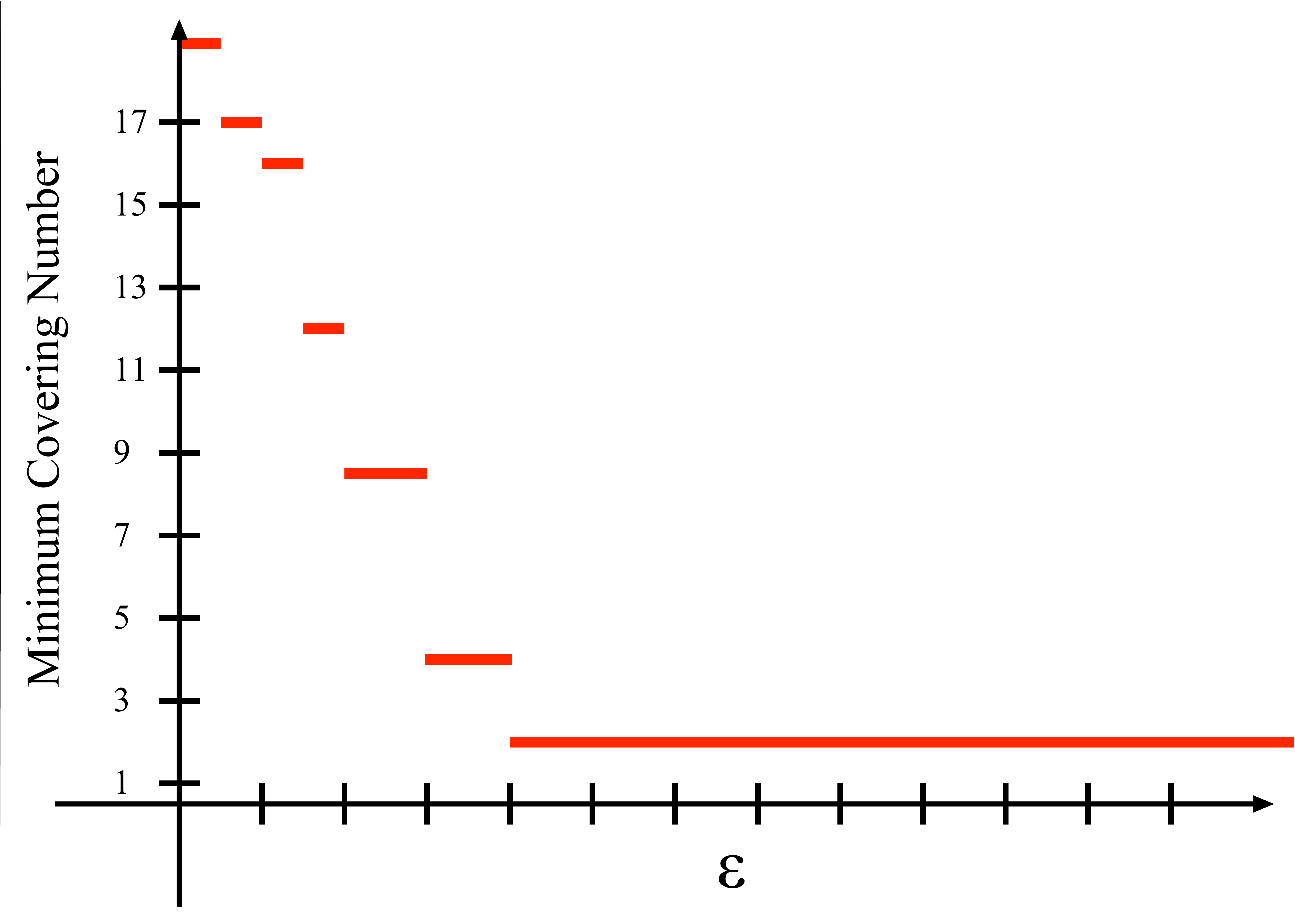}
        \caption{The minimum covering number of $S_1$ persists at 2 for more than 9 $\Delta\epsilon$'s.}
    \end{subfigure}   
    \caption{Comparison of persistent minimum covering numbers ($C_\epsilon$ as a function of $\epsilon$) of samples.}
    \label{fig:covdiag}
\end{figure}

We call the collection of the covering numbers which persists for large intervals of $\epsilon$, or, in a slight abuse of language, the collection of covering numbers for all $\epsilon$, the \emph{persistent minimum covering numbers}. Figure \ref{fig:covdiag} shows the plots of the persistent minimum covering numbers of the sets $S_1$ and $S_2$, from Figure \ref{fig:cov1}, as a function of $\epsilon$. These plots intuitively demonstrate the greater diversity of $S_2$, as $C_\epsilon(S_2)$ persists above 1 for a greater interval of $\epsilon$-values.

\section{Case Study} 

In the following, we consider three different synthetic data sets with
qualitatively different posterior structure: one with a single mode,
one with two modes, and one with an infinite number of modes (or
rather, a connected space of equally good solutions).  We consider
both a tiny version of the NMF problem, with only 4 or 5 observations
and 6 or 7 dimensions, as well as larger one with 50 observations and
500 dimensions.  We evaluate a number of sampling algorithms with 
various measures of ``diversity", 
running each algorithm for 10 repetitions of 10,000 samples.  To avoid
questions of whether samplers like Gibbs and HMC have mixed, we
initialize them at one of the modes/maximum likelihood solutions. 

\subsection{Synthetic Data Sets} 

\paragraph{Unique Solution}
Laurberg gives the following example,  $X = WA$, where the NMF solution is a unique equivalence class of factorizations, for the value $a = 0.3$. 
\begin{align}
W =\small \left(\begin{matrix}
a & 1 & 1& a & 0 & 0\\
1 & a & 0 & 0 & a & 1\\
0 & 0 & a & 1 & 1 & a
\end{matrix}\right), \;\;
A = W^T. \qquad \label{laurberg}
\end{align}
Under our Bayesian model, the posterior space of this NMF is unimodal.

\paragraph{Two Solutions}
Simply by setting the value of $a$ to be $0.5$ in the previous example, Laurberg shows that $X$ will have two distinct solutions. In particular, these solutions are related by a change of basis matrix $Q$, so the two factorization cones are in the same subspace. 
\begin{align}
W =\small \left(\begin{matrix}
0.5 & 1 & 1& 0.5 & 0 & 0\\
1 & 0.5 & 0 & 0 & 0.5 & 1\\
0 & 0 & 0.5 & 1 & 1 & 0.5
\end{matrix}\right), \;\;
A = W^T; \qquad 
Q = Q^{-1} =\frac{1}{3}\small \left(\begin{matrix}
-1 & 2& 2\\
2 & -1 & 2\\
2 & 2 & -1
\end{matrix}\right).
\end{align}
Under our Bayesian model, the posterior space of this NMF is bimodal.

\paragraph{Infinite Solutions}
Finally, for the following matrix,
\begin{align}
X & = \small\left(\begin{matrix}
1 & 1 & 1 & 0 & 0 & 0 &0 & 0 &0\\
0 & 0 & 0 & 1 & 1 & 1 &0 & 0 &0\\
0 & 0 & 0 & 0 & 0 & 0 &1 & 1 &1\\
1 & 0 & 0 & 1 & 0 & 0 &1 & 0 &0\\
0 & 1 & 0 & 0 & 1 & 0 &0 & 1 &0\\
0 & 0 & 1 & 0 & 0 & 1 &0 & 0 &1\\
1 & 1 & 1 & 1 & 1 & 1 &1 & 1 &1
\end{matrix}\right)
\end{align}
there are an \emph{infinite number} of nonnegative factorizations. For any $\delta \in [0,1]$, we have $X= A_\delta W$, where
\begin{align}
A_\delta = \small\left(\begin{matrix}
1 & 0 & 0 & 0 & 0 & 0\\
0 & 1& 0 & 0& 0 & 0\\
0 & 0& 1 & 0& 0 & 0\\
0 & 0& 0 & 1& 0 & 0\\
0 & 0& 0 & 0& 1 & 0\\
0 & 1& 0 & 0& 0 & 1\\
1-\delta & 1-\delta& 1-\delta & \delta& \delta &\delta
\end{matrix}\right),\quad
W = \small\left(\begin{matrix}
1 & 1 & 1 & 0 & 0 & 0 &0 & 0 &0\\
0 & 0 & 0 & 1 & 1 & 1 &0 & 0 &0\\
0 & 0 & 0 & 0 & 0 & 0 &1 & 1 &1\\
1 & 0 & 0 & 1 & 0 & 0 &1 & 0 &0\\
0 & 1 & 0 & 0 & 1 & 0 &0 & 1 &0\\
0 & 0 & 1 & 0 & 0 & 1 &0 & 0 &1\\
\end{matrix}\right)
\end{align} 
Under our Bayesian model, the posterior space of this NMF contains an infinite number of modes that form a connected subset of the parameter space.

\subsubsection{Larger Data Sets}
The above examples of data sets are valuable performing diagnostics on sampling algorithms and comparing diversity measures, because we know ground truth about the solution space for each example. The larger data sets we work with are simply high dimensional embeddings of these small data sets, chosen such that the desirable properties (such as the number of solutions) of the original data are preserved. To generate the larger data sets, we take a small $D \times N$ data $X$ and transform it using non-negative matrices $B_1$ and $B_2$ such that $X_{\text{large}} = B_1XB_2$. For an NMF data set, the factors transform in a simple manner: $A_{\text{large}} = B_1A$ and $W_{\text{large}}B_2$.  In our experiments, the larger data sets have dimensions $D = 500$ and $N = 50$ (while relatively small, we show that even with data sets of this size samplers rarely move very far).  

\subsection{Inference Approaches} 

In constructing our chain of factorization matrices, we add Gaussian noise, with standard deviation $\sigma$, to the data matrix $X$, for which we know the exact factorizations. We call the noisy data $\widetilde{X}$. 

\paragraph{Lower Bound: Single Mode}
To provide a reasonable lower bound on coverage, we establish the single-mode baseline, wherein we generate a set of samples that only explore one mode or region. For this baseline, we fix one known exact solution, $X = WA$; to generate samples consistent with the Gaussian noise model, we initialize an NMF algorithm at $(W, A)$ and iteratively generate factorizations for the data $X$ with added Gaussian noise. The procedure is as follows:  we initialize the multiplicative update algorithm of Lee and Seung \cite{Lee} with $(W, A)$; we then run the algorithm for the data $\widetilde{X} + E_i$, where $E_i$ is the noise matrix for the additional Gaussian noise added at the $i^{\text{th}}$ iteration.  In this fashion, we obtain variation in the entries of our factorization matrices (depending on the noise-level), but we know these variation is of a limited scale and that the chain contain samples from only one mode.  

\paragraph{Upper Bound: All Modes}
To provide a reasonable approximation of ideal coverage, we establish the all-modes baseline, wherein generates a set of samples that covers the entire solution space.  At each iteration in the chain, we randomly pick one known exact solution. When the solutions are discrete, we uniformly sample them. In the infinite solutions case, we pick $\delta_i$  independently from the uniform distribution over [0,1].  We then apply the same perturb-and-factor methodology as in the single mode baseline to produce a set of factors that approximate the modes of the parameter space. 


\paragraph{Gibbs}
Schmidt et al propose a Bayesian approach to NMF in which each basis
element $A_{d,r}$ and each weight $W_{r,n}$ is sampled independently
from an exponential prior. Combined with a Gaussian likelihood, the
updates for each element of $A$ and $W$ can be sampled element-wise
from a rectified normal distribution, which is proportional to the
product of a Gaussian and an exponential. 

We initialize with a sample from the \textit{All Modes} baseline so that no burn-in needed. We fix $\sigma$, the noise parameter in the Gaussian noise, as the same value used in generating $\widetilde{X}$.

\paragraph{Hamiltonian Monte Carlo} 

Hamiltonian Monte Carlo (HMC) is a Markov Chain Monte Carlo (MCMC)
procedure that simulates Hamiltonian dynamics over the target
distribution state space. The time-reversibility and volume preserving
properties of the evolution of Hamiltonian systems ensure detailed
balance \cite{Neal}. By incorporating gradient information, HMC
suppresses the random walk behavior which contributes to the
inefficiency of many MCMC methods.  Given a target density $\pi(q)$, we define a Hamiltonian of the form
\begin{align}
H(q, p) = -\log\pi_{\cH}(q) + \frac{1}{2} p^\top p
\end{align}
where $-\log\pi(q) $ is called the potential energy and the velocity,
$p$, is given by a certain tangent vector at $q$. We simulate the
dynamics with a discretized integrator, called the leap-frog
integrator.  Given an initial state $(q_0, p_0)$, a proposal state
$(q_n, p_n)$ is reached by a series of steps in the direction of
$\nabla_{q} \log \pi_{\cH}(q)$ \cite{Byrne}.

In the case of NMF, our target density, $\pi(W, A)$, is the log posterior distribution corresponding to our generative model. Since HMC have been demonstrated to be more successful than other MCMC methods in crossing multiple modes, for low-dimensional posterior spaces, one might expect HMC to out-perform Gibbs in finding multiple optima for NMF problems as well.

For our experiments, we implement HMC with adaptive step size and a fixed 100 leap count. We perform the HMC on the entry-wise log of the factor matrices in order to keep our solutions feasible. We initialize with a sample from the \textit{All Modes} baseline and allow an additional burn-in period of 200 iterations so that the we can adaptively find a reasonable step-size. 

\paragraph{Non-MCMC Baseline: (Filtered) Random Restarts}
We run Lin's projected gradient NMF algorithm \cite{lin2007projected} with random initializations. We expect this algorithm to explore multiple solutions when they exist because it is initialized at a random factor matrix each time. There are no asymptotic guarantees for exploring all solutions nor is there an underlying probabilistic framework.  However, random restarts provides a baseline of how much coverage we might achieve in practice if we did not have access to the true solution but were not constrained to a probabilistic MCMC framework.  

In practice, deterministic optimization algorithms such as the projected gradient algorithm  are prone to get caught in very poor local optima \cite{lin2007projected}.  Thus, we report metrics for all random restart outputs as well as the subset of random restarts results that achieve likelihoods comparable with the other samplers (that is, we exclude random restarts results from local optima corresponding to very poor solutions).  We call this second baseline \emph{filtered random restarts}. In our experiments, we allow for reconstruction errors in the filtered samples to be up to ten times the maximum reconstruction error in the Gibbs chain. 

\subsection{Evalution Metrics}
For our experiments, we report evaluation metrics measuring three key aspects of of sampling: quality of fit, sample independence and coverage of the parameter space.

\paragraph{Quality of Fit} 
We measure the quality of NMF solutions contained in a set of samples by computing the mean likelihood of the samples. Because methods like random restarts can produce samples from low-likelihood local optima that are widely distributed in the posterior space, the mean likelihood can be used to distinguish samples from multiple modes from those from multiple local optima. 

\paragraph{Sample Independence} 
We measure the independence of the samples in a set using integrated autocorrelation time. Although sample independence can be assessed in a number of different ways, we choose to use only integrated autocorrelation time since previous work \cite{cowles1996markov, brooks1998assessing, kass1998markov, brooks1998convergence} have shown IAT to be a reasonable representative of this class of metrics. 

\paragraph{Coverage} 
To asses the amount of coverage achieved by a set of samples, we compute the maximum pairwise distance, the mean pairwise distance, and the persistent minimum covering numbers. Furthermore, each metric of coverage is calculated with respect to \emph{both} maximum angle similarity and $\ell_1$-minimum matching distance.

The persistent minimum covering numbers are computed over a range of 100 $\epsilon$ values and for the first 1000 elements of each chain. For each $\epsilon$ value, the minimum covering number we record is the mean of $C_\epsilon$ over 10 repetitions of the same experiment. As the set cover problem is NP-hard, we employ a greedy algorithm to approximate the coverage \cite{chvatal1979greedy}.

\section{Results}\label{results}

\paragraph{Quality of Fit: Log-Likelihoods}
By design, the one mode and all modes sampling algorithms fit the data
well (see Table \ref{table:llsm} in Appendix \ref{sec:tabsm} and Table
\ref{table:lllg} in Appendix \ref{sec:tablg}).
Since the Gibbs sampler the HMC sampler are initialized at a known
mode, they also fit the data well while exploring the posterior space
around that mode. 

Random Restarts, on the other hand, has much lower likelihoods than
the samplers.  This result is expected (although perhaps not at this
scale). It appears that, even on small toy data sets, Random Restarts
is prone to being trapped in local optima around low-quality
solutions.  The presence of such poor factorizations in Random
Restarts motivates the construction of the Filtered Random Restarts
baseline, which contains the subset of Random Restart samples whose
reconstruction error in the Frobenius norm $\| \widetilde{X} - AW\|_F$
falls within a fixed threshold. In constructing the filtered samples,
we note that a large proportion of the Random Restarts samples from
the toy-sized data sets fall within our allowable reconstruction
criterion, while significantly fewer samples are allowed to filter
through when we consider the larger $500 \times 50$ data sets. In
particular, the worst performance of Random Restarts is observed in
the enlarged version of the bimodal data, wherein, within ten
repetitions, the number of sample satisfying our filtering criteria
ranged between 5 and 309 out of a total of 10,000 samples.

\paragraph{Sample Independence: IAT}
The MCMC algorithms show high degrees of autocorrelation rendering the
effective sample size of the 10,000 chain to be orders of magnitude
smaller (Table \ref{table:iatsm} in Appendix \ref{sec:tabsm} and Table
\ref{table:iatlg} in Appendix \ref{sec:tablg}). Even on toy data sets
with infinite number of connected solutions, the samples are highly
correlated. As expected, the non-MCMC samples are effectively
independent.

\paragraph{Coverage: Pairwise Distance and Minimum Covering Numbers}
We note again that the likelihood and sample independence does not
necessarily provide information about the region of the posterior that
is being explored in these chains. For example, in the infinite
solutions case, one can imagine a trajectory through the line of
solutions parameterized by $\delta$ that could generate a set of
samples with high autocorrelation but traverse along a large portion
of the posterior. On the other hand, samples could appear uncorrelated
when a chain is only making local but indpendent moves around a single
mode. To assess coverage, we study the maximum/mean pairwise distance
and persistent minimum covering numbers.

Both the pairwise distance measures (Table \ref{table:maxsm}, Table
\ref{table:maxlg}, Table \ref{table:meansm}, Table \ref{table:meanlg})
and the persistent covering numbers (Figures \ref{fig:cov1sm} -
\ref{fig:covinflg}) indicate that Random Restarts obtains the widest
range of solutions.  However, because some (or many, depending on the
data set) of the elements correspond to poor factorizations, they do
not represent samples from multiple modes of the posterior
space.\footnote{Occasionally, Random Restarts yields a solution with a
  column of zeros in the basis matrix $A$, this sort of degenerate
  sample gives rise to a maximum pairwise angle distance of 90 degree
  for the entire samples (Table \ref{table:maxsm}).} Thus, the
Filtered Random Restart samples reveals a better picture of the
movement of this sampling algorithm through the posterior. The
coverage metrics indicate that the Filtered Random Restart samples
still explore significantly more of the posterior than the MCMC
methods. In particular, in the toy sized bimodal data set, the
Filtered Random Restart samples appears to sample from both modes.

The persistence covering number plots of the All Modes samples
establish clear visual baselines for distinguishing sets containing
samples from multiple modes in the posterior space. Comparing the
plots of the Gibbs sampler and HMC against those of All Modes, we see
these asymptotically correct sampling mechanisms tend to explore
only a single mode (Figures \ref{fig:cov1sm} - \ref{fig:covinflg}).

\section{Conclusion}
We conclude from our experiments that the coverage metrics we introduce yield useful information regarding sampling behaviors that cannot be assessed using traditional MCMC diagnostics, such as the ability of the sampler to cover a large portion of the posterior space. We are able to evaluate and compare the exploratory behavior of chains through these metrics. Within the sampling algorithms we study, it appears that a filtered version of Random Restarts would show the best exploratory behavior while maintaining some quality of the factorization. In the data sets explored, the quality of the factorization deteriorates for Random Restarts as the scale of the data grows. 

Both MCMC approaches, Gibbs and HMC, produce factorizations of excellent quality but IAT shows that effective sample size is very small. Furthermore, our coverage analysis reveals that neither Gibbs nor HMC are able to explore the multiple modes in very small toy data sets. As the scale of the data set grows, we expect the Gibbs sampler to be more vulnerable to becoming trapped in a single mode due of the local nature of its updates. 

For the practitioner of NMF, these results first demonstrate that random restarts of some sort---such as running multiple chains---can still be important for covering the posterior space, even for very small problems.  More importantly, we argue for adding coverage analysis of solutions to the standard diagnostic process evaluating sampling-based approches, since solution set diversity is an unavoidable questing arising from the identifiability of NMF and other problems. These multiple solutions can lead to very different interpretations of the data and effect subsequent modeling decisions. We hope that the metrics introduced in this work to quantify the notion of sample diversity begins to fill the gap left by current performance measurement of sampling algorithms.


\newpage

\begin{appendices}

\section{Comparison of Metrics and Algorithms (Small Dataset)}\label{sec:tabsm}

\begin{table}[H]
\caption{Comparison of Performance of Sampling Method: Likelihood}
\label{table:llsm}
\begin{center}
\begin{tabular}{|cl|c|c|c|c|c|c|}
\hline
&&\parbox{1.7cm}{\centering One Mode}&\parbox{1.7cm}{\centering All Modes}&\parbox{1.7cm}{\centering ${}$\\Random\\Restarts\\}&\parbox{1.7cm}{\centering ${}$\\Random\\Restarts\\(filtered)\\}&\parbox{1.7cm}{\centering Gibbs}&\parbox{1.7cm}{\centering HMC}\\ 
\hline
\multirow{4}{*}{\parbox{1.5cm}{One\\Mode}} & \multirow{4}{*}{\parbox{1.5cm}{}}&-1.74e+01&-1.74e+01&-1.12e+03&-2.30e+02&-2.38e+01&-1.93e+01\\
 &&\tiny-1.84e+01, &\tiny-1.84e+01, &\tiny-1.22e+03, &\tiny-2.34e+02, &\tiny-2.52e+01, &\tiny-2.02e+01, \\
 &&\tiny-1.73e+01, &\tiny-1.73e+01, &\tiny-1.09e+03, &\tiny-2.32e+02, &\tiny-2.46e+01, &\tiny-1.92e+01, \\
 &&\tiny-1.65e+01&\tiny-1.65e+01&\tiny-1.02e+03&\tiny-2.27e+02&\tiny-2.34e+01&\tiny-1.88e+01\\
\hline
\multirow{4}{*}{\parbox{1.5cm}{Two\\Modes}} & \multirow{4}{*}{\parbox{1.5cm}{}}&-1.72e+01&-1.70e+01&-2.17e+03&-2.81e+02&-2.38e+01&-1.88e+01\\
 &&\tiny-1.80e+01, &\tiny-1.80e+01, &\tiny-2.24e+03, &\tiny-2.85e+02, &\tiny-2.51e+01, &\tiny-2.00e+01, \\
 &&\tiny-1.71e+01, &\tiny-1.69e+01, &\tiny-2.18e+03, &\tiny-2.79e+02, &\tiny-2.43e+01, &\tiny-1.93e+01, \\
 &&\tiny-1.62e+01&\tiny-1.63e+01&\tiny-2.07e+03&\tiny-2.77e+02&\tiny-2.32e+01&\tiny-1.75e+01\\
\hline
\multirow{4}{*}{\parbox{1.5cm}{Infinite\\Modes}} & \multirow{4}{*}{\parbox{1.5cm}{}}&-2.20e+01&-2.18e+01&-1.20e+05&-3.20e+02&-4.21e+01&-2.09e+01\\
 &&\tiny-2.37e+01, &\tiny-2.31e+01, &\tiny-1.22e+05, &\tiny-3.23e+02, &\tiny-4.30e+01, &\tiny-2.20e+01, \\
 &&\tiny-2.18e+01, &\tiny-2.15e+01, &\tiny-1.21e+05, &\tiny-3.21e+02, &\tiny-4.17e+01, &\tiny-2.05e+01, \\
 &&\tiny-1.94e+01&\tiny-2.01e+01&\tiny-1.20e+05&\tiny-3.19e+02&\tiny-4.15e+01&\tiny-1.92e+01\\
\hline
\end{tabular}
\end{center}
\end{table}

\begin{table}[H]
\caption{Comparison of Performance of Sampling Method: IAT}
\label{table:iatsm}
\begin{center}
\begin{tabular}{|cl|c|c|c|c|c|c|}
\hline
&&\parbox{1.7cm}{\centering One Mode}&\parbox{1.7cm}{\centering All Modes}&\parbox{1.7cm}{\centering ${}$\\Random\\Restarts\\}&\parbox{1.7cm}{\centering ${}$\\Random\\Restarts\\(filtered)\\}&\parbox{1.7cm}{\centering Gibbs}&\parbox{1.7cm}{\centering HMC}\\ 
\hline
\multirow{4}{*}{\parbox{1.5cm}{One\\Mode}} & \multirow{4}{*}{\parbox{1.5cm}{IAT}}&1.009&1.009&1.014&1.014&728.498&837.699\\
 &&\tiny1.01, &\tiny1.01, &\tiny1.01, &\tiny1.01, &\tiny584.8875, &\tiny822.665, \\
 &&\tiny1.01, &\tiny1.01, &\tiny1.015, &\tiny1.015, &\tiny727.725, &\tiny865.465, \\
 &&\tiny1.01&\tiny1.01&\tiny1.02&\tiny1.02&\tiny850.7075&\tiny940.6925\\
\hline
\multirow{4}{*}{\parbox{1.5cm}{Two\\Modes}} & \multirow{4}{*}{\parbox{1.5cm}{IAT}}&1.009&1.004&1.013&1.013&677.867&813.753\\
 &&\tiny1.01, &\tiny1.0, &\tiny1.01, &\tiny1.01, &\tiny643.415, &\tiny726.395, \\
 &&\tiny1.01, &\tiny1.0, &\tiny1.01, &\tiny1.01, &\tiny686.235, &\tiny831.19, \\
 &&\tiny1.01&\tiny1.01&\tiny1.0175&\tiny1.0175&\tiny719.4575&\tiny890.53\\
\hline
\multirow{4}{*}{\parbox{1.5cm}{Infinite\\Modes}} & \multirow{4}{*}{\parbox{1.5cm}{IAT}}&1.008&1.008&1.009&1.009&313.22&710.851\\
 &&\tiny1.01, &\tiny1.01, &\tiny1.01, &\tiny1.01, &\tiny292.9875, &\tiny681.7575, \\
 &&\tiny1.01, &\tiny1.01, &\tiny1.01, &\tiny1.01, &\tiny306.475, &\tiny703.79, \\
 &&\tiny1.01&\tiny1.01&\tiny1.01&\tiny1.01&\tiny320.755&\tiny724.35\\
\hline
\end{tabular}
\end{center}
\end{table}

\begin{table}[H]
\caption{Comparison of Performance of Sampling Method: Max}
\label{table:maxsm}
\begin{center}
\begin{tabular}{|cl|c|c|c|c|c|c|}
\hline
&&\parbox{1.7cm}{\centering One Mode}&\parbox{1.7cm}{\centering All Modes}&\parbox{1.7cm}{\centering ${}$\\Random\\Restarts\\}&\parbox{1.7cm}{\centering ${}$\\Random\\Restarts\\(filtered)\\}&\parbox{1.7cm}{\centering Gibbs}&\parbox{1.7cm}{\centering HMC}\\ 
\hline
\multirow{8}{*}{\parbox{1.5cm}{One\\Mode}} & \multirow{4}{*}{\parbox{1.5cm}{Angle\\Max}}&0.185&0.185&34.763&4.184&0.289&0.221\\
 &&\tiny0.18, &\tiny0.18, &\tiny28.395, &\tiny3.995, &\tiny0.28, &\tiny0.21, \\
 &&\tiny0.18, &\tiny0.18, &\tiny39.565, &\tiny4.055, &\tiny0.285, &\tiny0.215, \\
 &&\tiny0.19&\tiny0.19&\tiny39.6225&\tiny4.315&\tiny0.2975&\tiny0.2275\\
& \multirow{4}{*}{\parbox{1.5cm}{MM $\ell_1$\\Max}}&0.0&0.0&0.697&0.091&0.01&0.0\\
 &&\tiny0.0, &\tiny0.0, &\tiny0.575, &\tiny0.09, &\tiny0.01, &\tiny0.0, \\
 &&\tiny0.0, &\tiny0.0, &\tiny0.79, &\tiny0.09, &\tiny0.01, &\tiny0.0, \\
 &&\tiny0.0&\tiny0.0&\tiny0.79&\tiny0.09&\tiny0.01&\tiny0.0\\
\hline
\multirow{8}{*}{\parbox{1.5cm}{Two\\Modes}} & \multirow{4}{*}{\parbox{1.5cm}{Angle\\Max}}&0.152&36.965&38.132&37.895&0.278&0.219\\
 &&\tiny0.15, &\tiny36.9525, &\tiny37.8775, &\tiny37.7375, &\tiny0.27, &\tiny0.21, \\
 &&\tiny0.15, &\tiny36.96, &\tiny37.915, &\tiny37.89, &\tiny0.28, &\tiny0.21, \\
 &&\tiny0.15&\tiny36.9775&\tiny38.155&\tiny37.95&\tiny0.2875&\tiny0.2275\\
& \multirow{4}{*}{\parbox{1.5cm}{MM $\ell_1$\\Max}}&0.0&0.67&0.738&0.694&0.01&0.0\\
 &&\tiny0.0, &\tiny0.67, &\tiny0.73, &\tiny0.69, &\tiny0.01, &\tiny0.0, \\
 &&\tiny0.0, &\tiny0.67, &\tiny0.73, &\tiny0.69, &\tiny0.01, &\tiny0.0, \\
 &&\tiny0.0&\tiny0.67&\tiny0.7375&\tiny0.7&\tiny0.01&\tiny0.0\\
\hline
\multirow{8}{*}{\parbox{1.5cm}{Infinite\\Modes}} & \multirow{4}{*}{\parbox{1.5cm}{Angle\\Max}}&0.187&44.973&90.0&44.99&1.187&16.229\\
 &&\tiny0.18, &\tiny44.9375, &\tiny90.0, &\tiny44.9725, &\tiny0.985, &\tiny6.625, \\
 &&\tiny0.185, &\tiny44.975, &\tiny90.0, &\tiny44.99, &\tiny1.115, &\tiny17.605, \\
 &&\tiny0.1975&\tiny45.0075&\tiny90.0&\tiny44.9975&\tiny1.355&\tiny22.2925\\
& \multirow{4}{*}{\parbox{1.5cm}{MM $\ell_1$\\Max}}&0.009&1.0&2.0&1.004&0.032&0.383\\
 &&\tiny0.01, &\tiny1.0, &\tiny2.0, &\tiny1.0, &\tiny0.0225, &\tiny0.1825, \\
 &&\tiny0.01, &\tiny1.0, &\tiny2.0, &\tiny1.0, &\tiny0.03, &\tiny0.425, \\
 &&\tiny0.01&\tiny1.0&\tiny2.0&\tiny1.01&\tiny0.04&\tiny0.52\\
\hline
\end{tabular}
\end{center}
\end{table}

\begin{table}[H]
\caption{Comparison of Performance of Sampling Method: Mean}
\label{table:meansm}
\begin{center}
\begin{tabular}{|cl|c|c|c|c|c|c|}
\hline
&&\parbox{1.7cm}{\centering One Mode}&\parbox{1.7cm}{\centering All Modes}&\parbox{1.7cm}{\centering ${}$\\Random\\Restarts\\}&\parbox{1.7cm}{\centering ${}$\\Random\\Restarts\\(filtered)\\}&\parbox{1.7cm}{\centering Gibbs}&\parbox{1.7cm}{\centering HMC}\\ 
\hline
\multirow{8}{*}{\parbox{1.5cm}{One\\Mode}} & \multirow{4}{*}{\parbox{1.5cm}{Angle\\Mean}}&0.069&0.069&0.873&0.615&0.11&0.09\\
 &&\tiny0.07, &\tiny0.07, &\tiny0.75, &\tiny0.6, &\tiny0.11, &\tiny0.09, \\
 &&\tiny0.07, &\tiny0.07, &\tiny0.89, &\tiny0.61, &\tiny0.11, &\tiny0.09, \\
 &&\tiny0.07&\tiny0.07&\tiny0.985&\tiny0.6275&\tiny0.11&\tiny0.09\\
& \multirow{4}{*}{\parbox{1.5cm}{MM $\ell_1$\\Mean}}&0.0&0.0&0.019&0.01&0.0&0.0\\
 &&\tiny0.0, &\tiny0.0, &\tiny0.02, &\tiny0.01, &\tiny0.0, &\tiny0.0, \\
 &&\tiny0.0, &\tiny0.0, &\tiny0.02, &\tiny0.01, &\tiny0.0, &\tiny0.0, \\
 &&\tiny0.0&\tiny0.0&\tiny0.02&\tiny0.01&\tiny0.0&\tiny0.0\\
\hline
\multirow{8}{*}{\parbox{1.5cm}{Two\\Modes}} & \multirow{4}{*}{\parbox{1.5cm}{Angle\\Mean}}&0.06&18.467&18.61&18.563&0.102&0.089\\
 &&\tiny0.06, &\tiny18.4725, &\tiny18.6025, &\tiny18.5625, &\tiny0.1, &\tiny0.09, \\
 &&\tiny0.06, &\tiny18.48, &\tiny18.615, &\tiny18.575, &\tiny0.1, &\tiny0.09, \\
 &&\tiny0.06&\tiny18.4875&\tiny18.6275&\tiny18.58&\tiny0.1&\tiny0.09\\
& \multirow{4}{*}{\parbox{1.5cm}{MM $\ell_1$\\Mean}}&0.0&0.33&0.34&0.34&0.0&0.0\\
 &&\tiny0.0, &\tiny0.33, &\tiny0.34, &\tiny0.34, &\tiny0.0, &\tiny0.0, \\
 &&\tiny0.0, &\tiny0.33, &\tiny0.34, &\tiny0.34, &\tiny0.0, &\tiny0.0, \\
 &&\tiny0.0&\tiny0.33&\tiny0.34&\tiny0.34&\tiny0.0&\tiny0.0\\
\hline
\multirow{8}{*}{\parbox{1.5cm}{Infinite\\Modes}} & \multirow{4}{*}{\parbox{1.5cm}{Angle\\Mean}}&0.078&16.572&27.261&12.835&0.371&5.16\\
 &&\tiny0.08, &\tiny16.485, &\tiny26.6225, &\tiny12.3425, &\tiny0.2925, &\tiny1.65, \\
 &&\tiny0.08, &\tiny16.525, &\tiny27.33, &\tiny12.725, &\tiny0.315, &\tiny5.62, \\
 &&\tiny0.08&\tiny16.6775&\tiny27.69&\tiny13.3525&\tiny0.4425&\tiny6.1625\\
& \multirow{4}{*}{\parbox{1.5cm}{MM $\ell_1$\\Mean}}&0.0&0.373&0.566&0.284&0.011&0.123\\
 &&\tiny0.0, &\tiny0.37, &\tiny0.56, &\tiny0.27, &\tiny0.01, &\tiny0.0425, \\
 &&\tiny0.0, &\tiny0.37, &\tiny0.56, &\tiny0.28, &\tiny0.01, &\tiny0.135, \\
 &&\tiny0.0&\tiny0.3775&\tiny0.57&\tiny0.2975&\tiny0.01&\tiny0.1475\\
\hline
\end{tabular}
\end{center}
\end{table}

\newpage

\section{Comparison of Metrics and Algorithms (Large Dataset)}\label{sec:tablg}

\begin{table}[H]
\caption{Comparison of Performance of Sampling Method: Likelihood}
\label{table:lllg}
\begin{center}
\begin{tabular}{|cl|c|c|c|c|c|c|}
\hline
&&\parbox{1.7cm}{\centering One Mode}&\parbox{1.7cm}{\centering All Modes}&\parbox{1.7cm}{\centering ${}$\\Random\\Restarts\\}&\parbox{1.7cm}{\centering ${}$\\Random\\Restarts\\(filtered)\\}&\parbox{1.7cm}{\centering Gibbs}&\parbox{1.7cm}{\centering HMC}\\ 
\hline
\multirow{4}{*}{\parbox{1.5cm}{One\\Mode}} & \multirow{4}{*}{\parbox{1.5cm}{}}&-1.22e+04&-1.22e+04&-1.68e+06&-1.00e+06&-1.25e+04&-1.25e+04\\
 &&\tiny-1.23e+04, &\tiny-1.23e+04, &\tiny-1.82e+06, &\tiny-1.04e+06, &\tiny-1.26e+04, &\tiny-1.26e+04, \\
 &&\tiny-1.22e+04, &\tiny-1.22e+04, &\tiny-1.63e+06, &\tiny-9.99e+05, &\tiny-1.24e+04, &\tiny-1.24e+04, \\
 &&\tiny-1.22e+04&\tiny-1.22e+04&\tiny-1.50e+06&\tiny-9.57e+05&\tiny-1.24e+04&\tiny-1.24e+04\\
\hline
\multirow{4}{*}{\parbox{1.5cm}{Two\\Modes}} & \multirow{4}{*}{\parbox{1.5cm}{}}&-1.22e+04&-1.22e+04&-3.60e+06&-1.09e+06&-1.25e+04&-1.25e+04\\
 &&\tiny-1.23e+04, &\tiny-1.23e+04, &\tiny-3.74e+06, &\tiny-1.14e+06, &\tiny-1.26e+04, &\tiny-1.26e+04, \\
 &&\tiny-1.22e+04, &\tiny-1.22e+04, &\tiny-3.55e+06, &\tiny-1.10e+06, &\tiny-1.24e+04, &\tiny-1.24e+04, \\
 &&\tiny-1.22e+04&\tiny-1.22e+04&\tiny-3.21e+06&\tiny-1.04e+06&\tiny-1.24e+04&\tiny-1.24e+04\\
\hline
\multirow{4}{*}{\parbox{1.5cm}{Infinite\\Modes}} & \multirow{4}{*}{\parbox{1.5cm}{}}&-1.20e+04&-1.20e+04&-1.82e+06&-9.98e+05&-1.25e+04&-1.25e+04\\
 &&\tiny-1.20e+04, &\tiny-1.20e+04, &\tiny-1.94e+06, &\tiny-1.01e+06, &\tiny-1.25e+04, &\tiny-1.25e+04, \\
 &&\tiny-1.20e+04, &\tiny-1.20e+04, &\tiny-1.82e+06, &\tiny-9.93e+05, &\tiny-1.25e+04, &\tiny-1.25e+04, \\
 &&\tiny-1.20e+04&\tiny-1.19e+04&\tiny-1.65e+06&\tiny-9.73e+05&\tiny-1.24e+04&\tiny-1.24e+04\\
\hline
\end{tabular}
\end{center}
\end{table}

\begin{table}[H]
\caption{Comparison of Performance of Sampling Method: IAT}
\label{table:iatlg}
\begin{center}
\begin{tabular}{|cl|c|c|c|c|c|c|}
\hline
&&\parbox{1.7cm}{\centering One Mode}&\parbox{1.7cm}{\centering All Modes}&\parbox{1.7cm}{\centering ${}$\\Random\\Restarts\\}&\parbox{1.7cm}{\centering ${}$\\Random\\Restarts\\(filtered)\\}&\parbox{1.7cm}{\centering Gibbs}&\parbox{1.7cm}{\centering HMC}\\ 
\hline
\multirow{4}{*}{\parbox{1.5cm}{One\\Mode}} & \multirow{4}{*}{\parbox{1.5cm}{IAT}}&1.01&1.01&1.011&1.009&645.903&1173.558\\
 &&\tiny1.01, &\tiny1.01, &\tiny1.0025, &\tiny1.0025, &\tiny600.07, &\tiny1115.425, \\
 &&\tiny1.01, &\tiny1.01, &\tiny1.01, &\tiny1.01, &\tiny649.92, &\tiny1186.735, \\
 &&\tiny1.01&\tiny1.01&\tiny1.0175&\tiny1.01&\tiny694.5775&\tiny1286.695\\
\hline
\multirow{4}{*}{\parbox{1.5cm}{Two\\Modes}} & \multirow{4}{*}{\parbox{1.5cm}{IAT}}&1.01&1.011&1.012&1.082&524.915&1105.867\\
 &&\tiny1.01, &\tiny1.01, &\tiny1.0025, &\tiny1.0625, &\tiny434.6325, &\tiny1046.2725, \\
 &&\tiny1.01, &\tiny1.02, &\tiny1.01, &\tiny1.1, &\tiny530.575, &\tiny1124.91, \\
 &&\tiny1.01&\tiny1.02&\tiny1.02&\tiny1.1175&\tiny563.32&\tiny1180.215\\
\hline
\multirow{4}{*}{\parbox{1.5cm}{Infinite\\Modes}} & \multirow{4}{*}{\parbox{1.5cm}{IAT}}&1.01&1.008&1.005&1.014&1067.452&1077.456\\
 &&\tiny1.01, &\tiny1.0, &\tiny1.0, &\tiny1.0025, &\tiny1059.7475, &\tiny1071.205, \\
 &&\tiny1.01, &\tiny1.0, &\tiny1.005, &\tiny1.01, &\tiny1068.9, &\tiny1076.965, \\
 &&\tiny1.01&\tiny1.025&\tiny1.01&\tiny1.02&\tiny1077.79&\tiny1086.3875\\
\hline
\end{tabular}
\end{center}
\end{table}

\begin{table}[H]
\caption{Comparison of Performance of Sampling Method: Max}
\label{table:maxlg}
\begin{center}
\begin{tabular}{|cl|c|c|c|c|c|c|}
\hline
&&\parbox{1.7cm}{\centering One Mode}&\parbox{1.7cm}{\centering All Modes}&\parbox{1.7cm}{\centering ${}$\\Random\\Restarts\\}&\parbox{1.7cm}{\centering ${}$\\Random\\Restarts\\(filtered)\\}&\parbox{1.7cm}{\centering Gibbs}&\parbox{1.7cm}{\centering HMC}\\ 
\hline
\multirow{8}{*}{\parbox{1.5cm}{One\\Mode}} & \multirow{4}{*}{\parbox{1.5cm}{Angle\\Max}}&0.01&0.01&25.504&16.326&0.027&0.092\\
 &&\tiny0.01, &\tiny0.01, &\tiny22.93, &\tiny14.705, &\tiny0.0225, &\tiny0.08, \\
 &&\tiny0.01, &\tiny0.01, &\tiny25.68, &\tiny16.315, &\tiny0.03, &\tiny0.09, \\
 &&\tiny0.01&\tiny0.01&\tiny28.15&\tiny17.7075&\tiny0.03&\tiny0.0975\\
& \multirow{4}{*}{\parbox{1.5cm}{MM $\ell_1$\\Max}}&0.0&0.0&0.444&0.258&0.0&0.0\\
 &&\tiny0.0, &\tiny0.0, &\tiny0.4025, &\tiny0.24, &\tiny0.0, &\tiny0.0, \\
 &&\tiny0.0, &\tiny0.0, &\tiny0.44, &\tiny0.255, &\tiny0.0, &\tiny0.0, \\
 &&\tiny0.0&\tiny0.0&\tiny0.4825&\tiny0.285&\tiny0.0&\tiny0.0\\
\hline
\multirow{8}{*}{\parbox{1.5cm}{Two\\Modes}} & \multirow{4}{*}{\parbox{1.5cm}{Angle\\Max}}&0.01&10.779&22.131&5.084&0.02&0.072\\
 &&\tiny0.01, &\tiny10.6525, &\tiny17.6125, &\tiny3.405, &\tiny0.02, &\tiny0.07, \\
 &&\tiny0.01, &\tiny10.69, &\tiny23.08, &\tiny4.51, &\tiny0.02, &\tiny0.07, \\
 &&\tiny0.01&\tiny11.0125&\tiny26.05&\tiny6.765&\tiny0.02&\tiny0.0775\\
& \multirow{4}{*}{\parbox{1.5cm}{MM $\ell_1$\\Max}}&0.0&0.16&0.382&0.075&0.0&0.0\\
 &&\tiny0.0, &\tiny0.16, &\tiny0.2825, &\tiny0.05, &\tiny0.0, &\tiny0.0, \\
 &&\tiny0.0, &\tiny0.16, &\tiny0.4, &\tiny0.07, &\tiny0.0, &\tiny0.0, \\
 &&\tiny0.0&\tiny0.16&\tiny0.44&\tiny0.1&\tiny0.0&\tiny0.0\\
\hline
\multirow{8}{*}{\parbox{1.5cm}{Infinite\\Modes}} & \multirow{4}{*}{\parbox{1.5cm}{Angle\\Max}}&0.01&21.412&42.943&35.901&0.202&0.556\\
 &&\tiny0.01, &\tiny21.2, &\tiny41.9175, &\tiny33.4125, &\tiny0.19, &\tiny0.525, \\
 &&\tiny0.01, &\tiny21.41, &\tiny42.99, &\tiny35.68, &\tiny0.2, &\tiny0.555, \\
 &&\tiny0.01&\tiny21.6175&\tiny43.86&\tiny38.1225&\tiny0.225&\tiny0.5875\\
& \multirow{4}{*}{\parbox{1.5cm}{MM $\ell_1$\\Max}}&0.0&0.348&0.985&0.796&0.0&0.01\\
 &&\tiny0.0, &\tiny0.3425, &\tiny0.93, &\tiny0.7325, &\tiny0.0, &\tiny0.01, \\
 &&\tiny0.0, &\tiny0.35, &\tiny0.98, &\tiny0.785, &\tiny0.0, &\tiny0.01, \\
 &&\tiny0.0&\tiny0.35&\tiny1.0375&\tiny0.8375&\tiny0.0&\tiny0.01\\
\hline
\end{tabular}
\end{center}
\end{table}

\begin{table}[H]
\caption{Comparison of Performance of Sampling Method: Mean}
\label{table:meanlg}
\begin{center}
\begin{tabular}{|cl|c|c|c|c|c|c|}
\hline
&&\parbox{1.7cm}{\centering One Mode}&\parbox{1.7cm}{\centering All Modes}&\parbox{1.7cm}{\centering ${}$\\Random\\Restarts\\}&\parbox{1.7cm}{\centering ${}$\\Random\\Restarts\\(filtered)\\}&\parbox{1.7cm}{\centering Gibbs}&\parbox{1.7cm}{\centering HMC}\\ 
\hline
\multirow{8}{*}{\parbox{1.5cm}{One\\Mode}} & \multirow{4}{*}{\parbox{1.5cm}{Angle\\Mean}}&0.01&0.01&5.172&4.293&0.02&0.041\\
 &&\tiny0.01, &\tiny0.01, &\tiny4.7825, &\tiny4.15, &\tiny0.02, &\tiny0.04, \\
 &&\tiny0.01, &\tiny0.01, &\tiny5.075, &\tiny4.235, &\tiny0.02, &\tiny0.04, \\
 &&\tiny0.01&\tiny0.01&\tiny5.635&\tiny4.29&\tiny0.02&\tiny0.04\\
& \multirow{4}{*}{\parbox{1.5cm}{MM $\ell_1$\\Mean}}&0.0&0.0&0.077&0.062&0.0&0.0\\
 &&\tiny0.0, &\tiny0.0, &\tiny0.07, &\tiny0.06, &\tiny0.0, &\tiny0.0, \\
 &&\tiny0.0, &\tiny0.0, &\tiny0.075, &\tiny0.06, &\tiny0.0, &\tiny0.0, \\
 &&\tiny0.0&\tiny0.0&\tiny0.08&\tiny0.06&\tiny0.0&\tiny0.0\\
\hline
\multirow{8}{*}{\parbox{1.5cm}{Two\\Modes}} & \multirow{4}{*}{\parbox{1.5cm}{Angle\\Mean}}&0.01&5.38&3.467&2.457&0.02&0.033\\
 &&\tiny0.01, &\tiny5.33, &\tiny3.1325, &\tiny1.97, &\tiny0.02, &\tiny0.03, \\
 &&\tiny0.01, &\tiny5.35, &\tiny3.44, &\tiny2.75, &\tiny0.02, &\tiny0.03, \\
 &&\tiny0.01&\tiny5.49&\tiny3.8375&\tiny3.0&\tiny0.02&\tiny0.04\\
& \multirow{4}{*}{\parbox{1.5cm}{MM $\ell_1$\\Mean}}&0.0&0.08&0.051&0.036&0.0&0.0\\
 &&\tiny0.0, &\tiny0.08, &\tiny0.05, &\tiny0.03, &\tiny0.0, &\tiny0.0, \\
 &&\tiny0.0, &\tiny0.08, &\tiny0.05, &\tiny0.04, &\tiny0.0, &\tiny0.0, \\
 &&\tiny0.0&\tiny0.08&\tiny0.0575&\tiny0.04&\tiny0.0&\tiny0.0\\
\hline
\multirow{8}{*}{\parbox{1.5cm}{Infinite\\Modes}} & \multirow{4}{*}{\parbox{1.5cm}{Angle\\Mean}}&0.01&7.935&16.088&11.978&0.108&0.307\\
 &&\tiny0.01, &\tiny7.8125, &\tiny15.7175, &\tiny11.665, &\tiny0.1, &\tiny0.285, \\
 &&\tiny0.01, &\tiny7.905, &\tiny15.87, &\tiny11.885, &\tiny0.105, &\tiny0.305, \\
 &&\tiny0.01&\tiny8.0175&\tiny16.6375&\tiny12.16&\tiny0.1175&\tiny0.32\\
& \multirow{4}{*}{\parbox{1.5cm}{MM $\ell_1$\\Mean}}&0.0&0.129&0.258&0.186&0.0&0.001\\
 &&\tiny0.0, &\tiny0.13, &\tiny0.25, &\tiny0.18, &\tiny0.0, &\tiny0.0, \\
 &&\tiny0.0, &\tiny0.13, &\tiny0.26, &\tiny0.185, &\tiny0.0, &\tiny0.0, \\
 &&\tiny0.0&\tiny0.13&\tiny0.2675&\tiny0.19&\tiny0.0&\tiny0.0\\
\hline
\end{tabular}
\end{center}
\end{table}

\newpage

\section{Persistent Minimum Covering Number Plots (Small Dataset)}

\begin{figure}[H]
    \centering
    \begin{subfigure}[b]{0.45\textwidth}
        \centering
        \includegraphics[height=55mm]{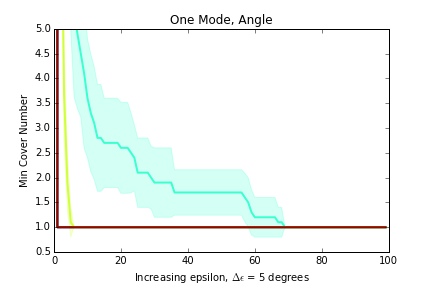}
        \caption{Similarity measure: Max Angle}
    \end{subfigure}    
    \begin{subfigure}[b]{0.45\textwidth}
        \centering
        \includegraphics[height=55mm]{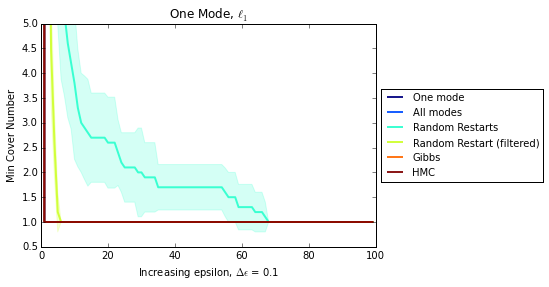}
        \caption{Similarity measure: MinMatch ($\ell_1$)}
    \end{subfigure}

    \caption{Unique Mode}\label{fig:cov1sm}
\end{figure}

\begin{figure}[H]
    \centering
    \begin{subfigure}[b]{0.45\textwidth}
        \centering
        \includegraphics[height=55mm]{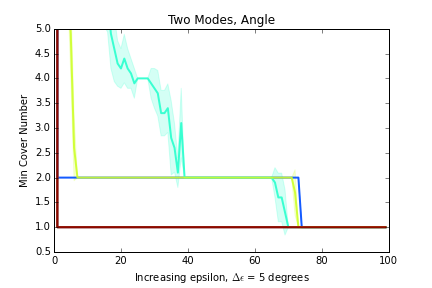}
        \caption{Similarity measure: Max Angle}
    \end{subfigure}    
    \begin{subfigure}[b]{0.45\textwidth}
        \centering
        \includegraphics[height=55mm]{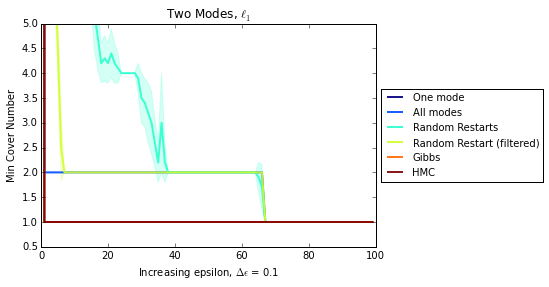}
        \caption{Similarity measure: MinMatch ($\ell_1$)}
    \end{subfigure}
    
    \caption{Two Modes}\label{fig:cov2sm}
\end{figure}

\begin{figure}[H]
    \centering
    \begin{subfigure}[b]{0.45\textwidth}
        \centering
        \includegraphics[height=55mm]{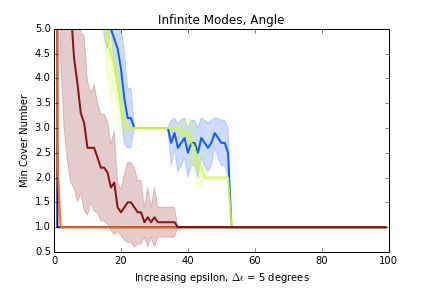}
        \caption{Similarity measure: Max Angle}
    \end{subfigure}    
    \begin{subfigure}[b]{0.45\textwidth}
        \centering
        \includegraphics[height=55mm]{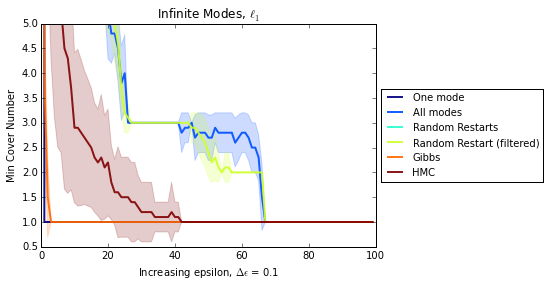}
        \caption{Similarity measure: MinMatch ($\ell_1$)}
    \end{subfigure}
    \caption{Infinite modes} \label{fig:covinfsm}
\end{figure}

\newpage

\section{Persistent Minimum Covering Number Plots (Large Dataset)}
\begin{figure}[H]
    \centering
    \begin{subfigure}[b]{0.45\textwidth}
        \centering
        \includegraphics[height=55mm]{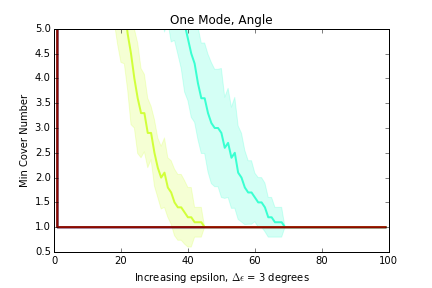}
        \caption{Similarity measure: Max Angle}
    \end{subfigure}    
    \begin{subfigure}[b]{0.45\textwidth}
        \centering
        \includegraphics[height=55mm]{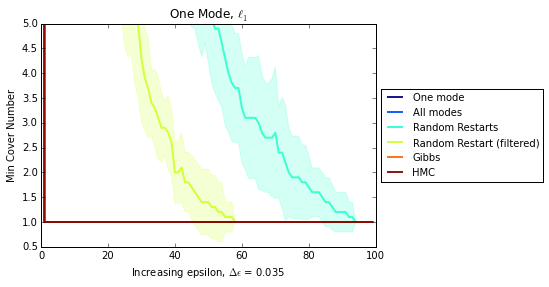}
        \caption{Similarity measure: MinMatch ($\ell_1$)} 
    \end{subfigure}

    \caption{Unique Mode}\label{fig:cov1lg}
\end{figure}

\begin{figure}[H]
    \centering
    \begin{subfigure}[b]{0.45\textwidth}
        \centering
        \includegraphics[height=55mm]{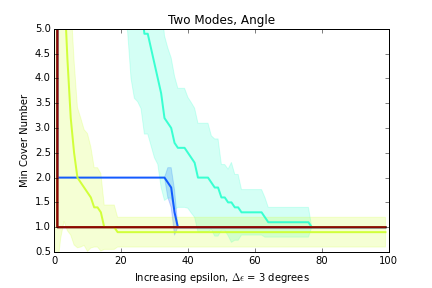}
        \caption{Similarity measure: Max Angle}
    \end{subfigure}    
    \begin{subfigure}[b]{0.45\textwidth}
        \centering
        \includegraphics[height=55mm]{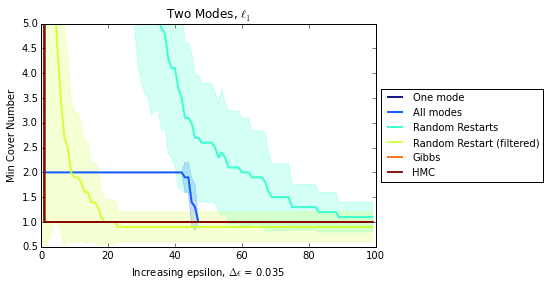}
        \caption{Similarity measure: MinMatch ($\ell_1$)}
    \end{subfigure}
    
    \caption{Two Modes}\label{fig:cov2lg}
\end{figure}

\begin{figure}[H]
    \centering
    \begin{subfigure}[b]{0.45\textwidth}
        \centering
        \includegraphics[height=55mm]{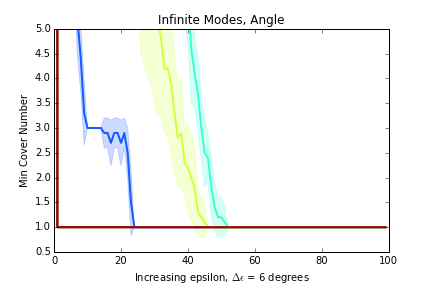}
        \caption{Similarity measure: Max Angle}
    \end{subfigure}    
    \begin{subfigure}[b]{0.45\textwidth}
        \centering
        \includegraphics[height=55mm]{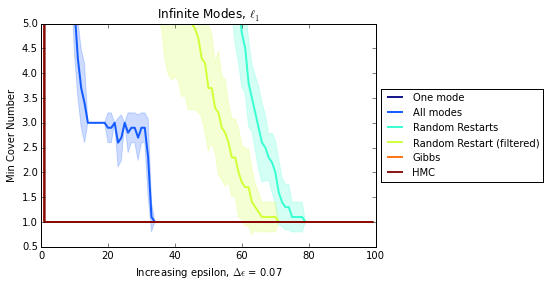}
        \caption{Similarity measure: MinMatch ($\ell_1$)}
    \end{subfigure}
    \caption{Infinite modes}\label{fig:covinflg}
\end{figure}
\end{appendices}

\bibliographystyle{IEEEtranN}
\bibliography{posterior_mvmt_metric}

\end{document}